\documentclass{article}
\usepackage{microtype}
\usepackage{graphicx}
\usepackage{booktabs} 
\usepackage[table,xcdraw,x11names]{xcolor}
\usepackage{makecell}
\usepackage{array}
\usepackage{colortbl}             
\usepackage{graphicx}
\usepackage{subcaption}
\usepackage{float}
\usepackage{multirow}
\usepackage{tabularx}
\usepackage[normalem]{ulem}
\usepackage{hyperref}

\usepackage[accepted]{icml2025}

\usepackage{amsmath}
\usepackage{amssymb}
\usepackage{mathtools}
\usepackage{amsthm}
\usepackage{pgfplots}
\usepackage{pgfplotstable}
\pgfplotsset{compat=1.18}
\usepackage[capitalize,noabbrev]{cleveref}

\theoremstyle{plain}

\theoremstyle{definition}

\theoremstyle{remark}

\usepackage{xspace}
\newcommand*{\eg}{\emph{e.g.},\@\xspace}
\newcommand*{\ie}{\emph{i.e.},\@\xspace}

\newcommand*{\vs}{\emph{v.s.}\@\xspace}

\usepackage[textsize=tiny]{todonotes}
\usetikzlibrary{patterns}
\definecolor{LGray}{gray}{0.97}

\definecolor{IllinoisOrange}{HTML}{FF5E00}
\definecolor{IllinoisBlue}{HTML}{406080}

\icmltitlerunning{Confidence Elicitation in Embodied Agents}

\begin{document}

\twocolumn[

\icmltitle{Uncertainty in Action: Confidence Elicitation in Embodied Agents}

\icmlsetsymbol{equal}{*}

\begin{icmlauthorlist}
    \icmlauthor{Tianjiao Yu,}{}
    \icmlauthor{Vedant Shah,}{}
    \icmlauthor{Muntasir Wahed,}{}
    \icmlauthor{Kiet A. Nguyen,}{}
    \icmlauthor{Adheesh Juvekar}{}\\
    \icmlauthor{Tal August,}{}
    \icmlauthor{Ismini Lourentzou}{}
\end{icmlauthorlist}
\vspace{-0.2cm}
\begin{center}
\textcolor{IllinoisOrange}{University of Illinois Urbana-Champaign} \\
{\tt\small \{ty41,vrshah4,mwahed2,kietan2,adheesh2,taugust,lourent2\}@illinois.edu}\\
 \url{https://plan-lab.github.io/ece}
\end{center}

\icmlcorrespondingauthor{Tianjiao Yu}{ty41@illinois.edu}
\icmlkeywords{Embodied Agents, Uncertainty in Embodied AI, Verbalized Uncertainty}

\vskip 0.3in
]
\thispagestyle{plain} 
\renewcommand{\footnoterule}{\hrule width 0.5\textwidth height 0.4pt \vspace{3pt}} 
\footnotetext{\small *Preprint. Work in progress.} 


\begin{abstract}
Expressing confidence is challenging for embodied agents navigating dynamic multimodal environments, where uncertainty arises from both perception and decision-making processes.
We present the first work investigating embodied confidence elicitation in open-ended multimodal environments. We introduce Elicitation Policies, which structure confidence assessment across inductive, deductive, and abductive reasoning, along with Execution Policies, which enhance confidence calibration through scenario reinterpretation, action sampling, and hypothetical reasoning.
Evaluating agents in calibration and failure prediction tasks within the Minecraft environment, we show that structured reasoning approaches, such as Chain-of-Thoughts, improve confidence calibration. However, our findings also reveal persistent challenges in distinguishing uncertainty, particularly under abductive settings, underscoring the need for more sophisticated embodied confidence elicitation methods.
\end{abstract}
\section{Introduction}

In complex embodied environments, success depends not only on what an agent knows but also on how well it understands and communicates uncertainty. 
Whether navigating a cluttered space, interacting with objects, or planning long-term strategies, eliciting confidence is pivotal as agents must interpret and interact with dynamic settings in real-time while managing uncertainty from both perception and decision-making processes~\cite{ren2023robots, liang2024introspective}. 
For humans, this instinctive ability to express and calibrate uncertainty is fundamental to decision-making and social interaction. 
As AI systems are increasingly deployed in high-stakes contexts such as autonomous driving or healthcare, they must also acquire this crucial skill.

Specifically, accurate confidence elicitation from AI systems provides critical insights for risk assessment, error mitigation, and system reliability in decision-making~\cite{kuleshov2022calibrated,clark2015surfing, yildirim2019leveraging}. This is particularly important in open-ended reasoning tasks, where models may generate outputs that are semantically plausible but factually incorrect, a phenomenon commonly referred to as hallucination~\cite{xiao2021hallucination}. However, confidence elicitation in embodied AI is particularly challenging.
For instance, in open-ended environments such as Minecraft, an agent may misinterpret visual cues due to limited viewpoints or struggle to determine the correct action sequence to achieve complex goals (\eg obtaining a diamond). These illustrate the broader difficulties in eliciting confidence in embodied environments, where agents must navigate uncertainty at multiple levels.

\begin{figure}[t!]
  \centering
    \includegraphics[width=0.85\columnwidth]{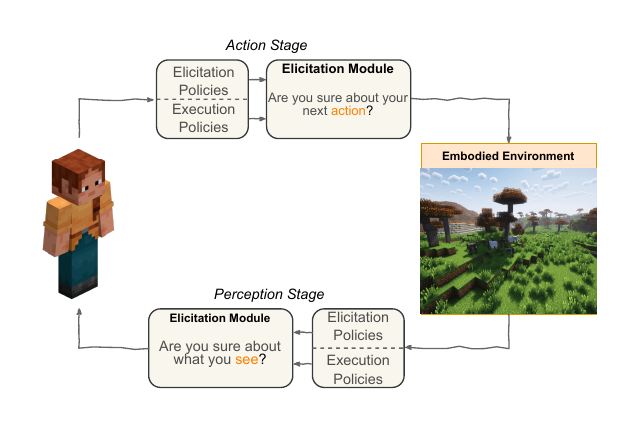}
    \vspace{-0.3cm}
  \caption{\textbf{Embodied Confidence Estimation Framework} consisting of \textit{Elicitation Policies} and \textit{Execution Policies}, which jointly enable an agent to assess and express its confidence. Elicitation Modules prompt the agent to evaluate uncertainty in what it sees and does, while \textit{Execution Policies} refine confidence calibration by expanding the agent's reasoning space (See \S\ref{sec:method} for details).}
  \label{fig:framework}
  \vspace{-0.3cm}
\end{figure}

Confidence elicitation in open-ended embodied environments faces several challenges, including: 1) Multimodal understanding, where the agent must assess uncertainty from inputs across different interconnected modalities. 
2) Granularity of confidence estimation, where the agent evaluates confidence not only in performing specific actions (\eg ``I am 90\% confident I can collect some wood") but also in understanding high-level tasks or goals (\eg ``I am 70\% confident I craft a wooden table"). 
3) Interactive dependencies, where the agent's actions directly influence the environment, which in turn affects subsequent decisions, requiring ongoing adjustments to confidence estimates as tasks progress. 4) Finally, while state-of-the-art embodied agents leverage proprietary Large Language Models (LLMs) and Vision-Language Models (VLMs) for their strong multimodal understanding and reasoning capabilities ~\cite{wang2023voyager, qin2024mp5, zhu2023ghost}, these often lack access to internal token likelihoods or probabilistic outputs, making traditional confidence estimation methods ineffective ~\cite{kumar2023conformal, chen2024knowing}.

To address these challenges, we present the first systematic approach that enables LLM/VLM-powered embodied agents to assess and articulate their confidence across multimodal inputs, multiple granularities, and dynamic embodied environments. Our contributions are as follows:
\textbf{(1)} We propose a framework for embodied verbalized confidence elicitation in multimodal open-ended environments. \textbf{(2)} As illustrated in Figure \ref{fig:framework}, we introduce Elicitation and Execution Policies to enhance confidence estimation in embodied settings. \textbf{Elicitation Policies} target different types of uncertainties arising from inductive, deductive, and abductive reasoning, while also facilitating multi-granular confidence estimation, allowing agents to assess uncertainty at both perception and action stages. \textbf{Execution Policies} improve robust elicitation across diverse scenarios, plans, and actions while tackling interactive dependencies by incorporating additional information about the environment and expanding potential action trajectories. \textbf{(3)} We provide the first structured analysis of embodied uncertainty and identify effective methods for improving confidence calibration and failure prediction, while also pinpointing persistent challenges. 

The following are key observations from our analysis:
\\\textbf{(1) Elicitation Policies are Effective But Vary by Context:} While all proposed elicitation policies improve confidence calibration and failure prediction, their effectiveness varies based on task complexity and uncertainty type, highlighting the need for adaptive strategies that align with the embodied agent's reasoning process and environment demands.\\
\textbf{(2) Execution Policies Amplify Reliable Embodied Confidence Elicitation:} Execution policies enhance the robustness of elicited confidence as they expand the range of available actions and scenario interpretations, enabling agents to assess their confidence levels more effectively based on a broader set of potential outcomes.\\
\textbf{(3) Model Differences Persist}: While all models benefit from the proposed policies, differences in their inherent reasoning and representation capabilities lead to significant variability in confidence calibration and task success rates, highlighting the importance of tailoring elicitation and execution strategies to each model's strengths and limitations.

\section{Related Works}

\begin{figure*}[ht!]
  \centering
  \includegraphics[width=0.98\textwidth]{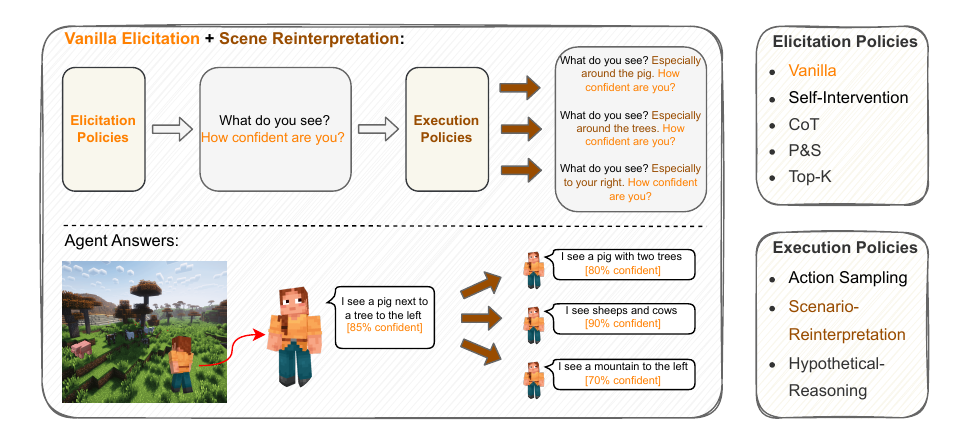}
  \vspace{-0.7cm}
  \caption{\textbf{Embodied Confidence Elicitation.} \textit{Elicitation Policies} (\S\ref{sec:Elicitation_Policies}) enable agents to express uncertainty, while \textit{Execution Policies} (\S\ref{sec:Execution_Policies}) refine and expand confidence assessment through scenario reinterpretation, action sampling, and hypothetical reasoning. Together, they enhance confidence calibration in embodied agents. The \textcolor{orange}{orange text} represents the vanilla elicitation policy, which incorporates the vanilla confidence prompt (described in Table \ref{tab:prompts}) into the original instruction. The brown arrows \protect\raisebox{-0.2\height}{\includegraphics[width=0.5cm]{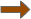}} denote the Scenario-Reinterpretation execution policy, prompting the agent to generate additional scene insights.}
  \label{fig:policy_example}
  \vspace{-0.3cm}
\end{figure*}

\textbf{Confidence Elicitation.}
Confidence elicitation for traditional machine learning is well-studied \cite{abdar2021review, gawlikowski2022survey}. One stream of work focuses on unsupervised methods leveraging entropy \cite{malinin2021uncertainty}, graph semantic parsing \cite{lin2022towards}, semantic features \cite{kuhn2023semantic, farquhar2024detecting}, and logit or hidden state information \cite{su2024unsupervised, chen2024inside} to craft uncertainty metrics. Another explores conformal prediction for tasks like part-of-speech tagging \cite{dey2022conformal}, paraphrase detection \cite{giovannotti2021transformer}, and fact verification \cite{fisch2020efficient}, offering statistically robust coverage guarantees \cite{kumar2023conformal, ye2024benchmarking}.

However, these solutions often require full model access, making them less applicable to black-box language models, which are increasingly prevalent in real-world applications \cite{openai2024gpt4technicalreport, touvron2023llamaopenefficientfoundation}. Additionally, their free-form nature of outputs further complicates the application of traditional methods. As a result, alternative approaches have been proposed, including estimating uncertainty by directly querying models for confidence scores after generating responses \cite{xiong2024can, kadavath2022languagemodelsmostlyknow, lin2022teachingmodelsexpressuncertainty, mielke2022reducingconversationalagentsoverconfidence, chen2023quantifyinguncertaintyanswerslanguage}. Despite these advancements, existing methods are not designed for embodied tasks, where confidence elicitation must address the challenges of multimodal perception, hierarchical reasoning and planning across various open-ended tasks, as well as non-deterministic interactions.

\textbf{LLM-based Embodied Agents.}
With the advent of language models, leveraging their reasoning and planning abilities to empower embodied agents has become quintessential \cite{huang2022inner, yao2023reactsynergizingreasoningacting, chen2023fireactlanguageagentfinetuning, zhang2024agentohanadesignunifieddata, shinn2024reflexion, christianos2023panguagentfinetunablegeneralistagent}. In the meantime, Minecraft’s open-ended nature with its adaptable mechanics and varied challenges, makes it a compelling benchmark for embedding reasoning and planning capabilities into language model-driven embodied agents \cite{wang2023voyager, wang2023describe, zhu2023ghost}.
Recent works leverage pre-trained language models to control agents by generating continuous operation instructions or executable policies. For example, some approaches \cite{zhu2023ghost, wang2023describe} directly utilize scene data from simulation platforms like MineDojo \cite{fan2022minedojo} and MineRL \cite{guss2019minerl}, while others \cite{qin2024mp5} rely on Vision-Language Models (VLMs) for perception.

However, because language models are used in various roles—such as planners, critics, or perceivers—errors and inaccuracies often arise at different process stages \cite{guo2024embodied, driess2023palmeembodiedmultimodallanguage}. These challenges underscore the need for frameworks capable of systematically identifying and localizing sources of uncertainty, which we aim to address by designing a unified approach that enhances reliability and robustness in embodied agents.

\textbf{Uncertainty in Embodied Models.}
Uncertainty estimation is well explored in robot learning and reinforcement learning ~\cite{wang2021online, ghasemipour2022so, he2023robust, huang2019bootstrap, jin2023neu}, but remains a challenge for language models ~\cite{tian2023just, groot2024overconfidence, zhang2024vl}. While recent efforts have sought to quantify and mitigate uncertainty~\cite{sagar2024failures, tian2022vibus}, the problem is further compounded in embodied AI settings, where agents must reason and act in dynamic multimodal environments \cite{ren2024explore,shen2023learning}. Our work introduces a structured approach to verbalized confidence elicitation in embodied open-ended multimodal environments to enable agents to express uncertainty and adapt to complex real-world interactions.

\section{Method}\label{sec:method}

\subsection{Problem Formulation \& Framework Design}
Let $\mathcal{E}$ denote the embodied environment, characterized by multimodal sensory inputs $\mathcal{I}\!=\!\{\mathcal{I}_v, \mathcal{I}_t\}$, where $\mathcal{I}_v$ represents visual observations and $\mathcal{I}_t$ represents task instructions and other types of language-based guidance.  For a given task $\mathcal{T}$, the agent operates under a policy $\pi: \mathcal{I} \rightarrow \mathcal{A}$ that maps input $\mathcal{I}$ to actions $\mathcal{A}$. The task of embodied confidence elicitation is to enable agents to estimate and articulate a confidence score $c \in [0, 1]$, representing their belief in the correctness of their perception and subsequent actions. 

The challenge lies in systematically identifying, quantifying, and articulating uncertainty as the agent interacts with its environment and executes tasks. This requires not only detecting uncertain aspects of the agent’s perception, reasoning, or actions but also ensuring that confidence estimates are refined and reliable under dynamic multimodal conditions. 
To address this, we propose an embodied confidence estimation framework centered around \textbf{Elicitation Modules} that facilitates confidence elicitation at two critical points of interaction between the agent and its environment:
\noindent \textbf{Perception Stage}, where the agent processes sensory input from the environment and assesses its confidence in what it perceives before engaging in reasoning or planning.
\noindent \textbf{Action Stage}, which evaluates the agent's confidence after reasoning, just before executing an action.

Each Elicitation Module operates under a specific \textbf{Elicitation Policy} (\S\ref{sec:Elicitation_Policies}), which defines \textit{what} type of uncertainty is being expressed, focusing on quantifying confidence in the agent's perception, reasoning, or action planning. Additionally, an \textbf{Execution Policy} (\S\ref{sec:Execution_Policies}) determines \textit{how} to collect and refine confidence, ensuring robust and adaptive estimates in complex, dynamic environments. An overview of the overall proposed method is shown in Figure \ref{fig:policy_example}.

\begin{table*}[t!]
    \centering
    \small
    \begin{tabularx}{0.96\textwidth}{c X}
        \hline
        \textbf{Method} & \textbf{Prompt}\\
        \hline
         \multirow{6}{*}{\centering Vanilla} & Read the task (e.g., collect wood, build a shelter), provide your answer, and explain \textcolor{OrangeRed1}{how confident you are in perceiving the environment accurately to complete the task} (e.g., recognizing resources, locating structures, identifying threats). \newline Read the task given, provide your answer, and explain \textcolor{Blue4}{how confident you are in planning and executing the actions needed to achieve the goal} (e.g., gathering materials, crafting tools, building a structure).\\
        \hline
        \multirow{4}{*}{\parbox{3cm}{\centering Vanilla \\ + \\ Self-Intervention}} & Task: [...], \textcolor{OrangeRed1}{Perceived Situation}: [...] Q: \textcolor{OrangeRed1}{How confident you are in perceiving the environment accurately to complete the task?} 
        \newline Task: [...], \textcolor{Blue4}{Planned Action}: [...] Q: \textcolor{Blue4}{How confident you are in planning and executing the actions needed to achieve the goal?}\\
        \hline
        \multirow{5}{*}{\centering\parbox{3cm}{\centering Chain-of-Thought \\ (CoT)}} & Read the task, \textcolor{OrangeRed1}{analyze step by step what you perceive in the environment} (e.g., observe surroundings, identify items), provide your answer, and evaluate your confidence \textcolor{OrangeRed1}{based on the clarity and quality of the environment observations}.
        \newline Read the task, \textcolor{Blue4}{analyze step by step how to complete the task}, provide your answer, and evaluate your confidence in \textcolor{Blue4}{successfully planning and executing each action needed to achieve the goal}.\\
        \hline
        \multirow{5}{*}{\parbox{3cm}{\centering Plan \& Solve \\ (P\&S)}} & Analyze the task, \textcolor{OrangeRed1}{devise a systematic approach to perceive your environment effectively.} (e.g., locating resources, identifying obstacles), and evaluate your confidence \textcolor{OrangeRed1}{based on how well you perceive the environment}.
        \newline Analyze the task, \textcolor{Blue4}{devise a plan of actions needed to complete it}, then evaluate your confidence \textcolor{Blue4}{in executing each action and achieving the desired outcome}.\\ 
        \hline
        \multirow{4}{*}{\centering Top-K} & Provide your \textcolor{OrangeRed1}{K best descriptions of your perceptions of the environment and the probability that each is correct (0\% to 100\%)}.
        \newline Provide your \textcolor{Blue4}{K best plans of the possible actions to take and the probability that each will succeed (0\% to 100\%)}.\\
        \hline
    \end{tabularx}
    \vspace{-0.2cm}
    \caption{\textbf{Prompts for Different Elicitation Policies} in generalist embodied Minecraft agents. \textcolor{OrangeRed1}{Orange text} indicates prompts focused on perception, while \textcolor{Blue4}{blue text} highlights prompts centered on action and planning.}
    \label{tab:prompts}
    \vspace{-0.3cm}
\end{table*}
\subsection{Elicitation Policies}
\label{sec:Elicitation_Policies}
Our confidence Elicitation Policies are designed to address distinct types of inferential uncertainty that embodied agents encounter in open-world long-horizon tasks. As these agents actively reason to determine their next actions, we draw inspiration from rich studies on reasoning in language models \cite{huang2023reasoninglargelanguagemodels} and introduce five prompt instructions, comprising two general-purpose methods and three tailored to inductive, deductive, and abductive reasoning settings (Appendix \ref{appendix:uncertainty_definition}). These prompts ask the agent to verbalize its confidence levels and systematically refine its uncertainty. Table \ref{tab:prompts} provides an overview of elicitation policy types with corresponding examples.

$\diamond$ \textbf{Vanilla.} Leveraging the inherent capability of language models ~\cite{brown2020language, wei2022emergent}, the Vanilla method directly queries the agent's confidence without additional structure or intervention. Vanilla serves as a baseline for comparison, relying solely on the agent's built-in capacity for confidence elicitation and self-assessment.

$\diamond$ \textbf{Self-Intervention}. Humans naturally benefit from revisiting their decisions with a fresh perspective, often uncovering insights or errors they initially overlooked. Inspired by this, the self-intervention method separates answer generation from evaluation. In one session, the model generates an answer; in another, it revisits the question and its response to assess its accuracy. This independent second pass mitigates confirmation bias and overconfidence, encouraging critical self-reflection and producing more reliable evaluations. 

$\diamond$ \textbf{Chain-of-Thought (CoT)}. To address uncertainty in inductive reasoning settings, where the agent must identify patterns and infer relationships from observations, we employ zero-shot Chain-of-Thought (CoT) reasoning \cite{wei2022chain}. 
By decomposing tasks into incremental steps, CoT enhances both interpretability and confidence calibration, allowing agents to reassess uncertainty at each step. 

$\diamond$ \textbf{Plan \& Solve (P\&S)}. Despite the success of CoT, it often suffers from semantic misunderstandings and missing step errors, particularly when applying general principles to specific cases. These failures stem from uncertainty in deductive reasoning, where the agent is unsure about the correct instantiation of abstract rules or whether a logical step is valid in a given context. P\&S ~\cite{wang2023plan} mitigates this by explicitly separating planning from execution, prompting the agent to construct a structured reasoning blueprint before solving the problem step by step. 

$\diamond$ \textbf{Top-K}. To address uncertainty in {abductive} reasoning, where multiple plausible explanations may fit the observed data, the Top-K method prompts the agent to generate its top K answers, each with an associated confidence level. This encourages the agent to consider and distribute its attention across several possible outcomes. By ranking responses rather than a single definitive answer, Top-K provides a balanced and comprehensive representation of abductive uncertainty across multiple plausible interpretations.

\subsection{Execution Policies}
\label{sec:Execution_Policies}
In embodied contexts, planning is a key factor in task success, requiring the agent to assess its confidence in executing action sequences effectively. Dynamic environments introduce unpredictable factors in action outcomes, which makes it important for the agent to not only consider its primary course of action but also to evaluate and communicate its confidence in alternative actions. By analyzing variance across potential actions rollouts, the agent can better quantify uncertainty and anticipate divergent outcomes. To address this, we introduce a set of policies that generate additional observations and diverse action trajectories, promoting robust confidence assessment:

$\circlearrowright$ \textbf{Action Sampling}: The agent can generate multiple possible actions by sampling from a learned policy distribution over the action space, conditioned on the current state and task objectives. By doing so, the agent can explore multiple actions, evaluate different outcomes, and assess which is most likely to succeed based on its perception.

$\circlearrowright$ \textbf{Scenario Reinterpretation}: The agent can be prompted to reinterpret the same scenario from different perspectives. For example, it could focus on a particular object, re-evaluate environmental obstacles, or re-assess the proximity of targets. This enables the agent to propose different courses of action by gathering and redirecting its attention to relevant environmental information.

$\circlearrowright$ \textbf{Hypothetical Reasoning}: The agent can be prompted with hypothetical or counterfactual scenarios (\eg ``What if the object in front were not an obstacle?"). By simulating these hypotheticals, the agent can explore how its actions would change and assess confidence in its original plan. This helps to gauge how flexible the agent's decision-making process is when confronted with uncertainty or alternative interpretations of the environment.

Figure \ref{fig:policy_example} provides an overview and examples of Elicitation and Execution Policies. During task-solving, agents rely on these execution policies to gather additional information about the environment and potential action trajectories, which they incorporate into further confidence elicitation.

\section{Experiment Setup}
\textbf{Environment \& Task Setting.} Minecraft has emerged as a popular benchmark for embodied AI research due to its open-ended environment, with diverse terrains, resources, and open-ended goals, making it an ideal testbed for embodied agents that perform hierarchical reasoning and long-term planning~\cite{johnson2016malmo, guss2019minerl, hafner2023mastering, nottingham2023embodied,lin2023mcu, qin2024mp5}. Building on this foundation, we define $30$ tasks evenly distributed across three difficulty levels: easy, medium, and hard, based on the complexity of reasoning steps and the amount of contextual information required. Detailed task descriptions can be found in Appendix \ref{appendix:task_setting}.

Easy tasks typically involve basic interactions with a single environmental element (\eg locating a pig or observing the weather). 
Medium tasks require combining perception and reasoning over multiple elements, while hard tasks increase dependency on sequential reasoning and include complex challenges like the \textit{Diamond Challenge}, which requires long-term planning and multi-step execution. Following prior work~\cite{guss2019minerl}, the maximum episode length is set to 6000 steps. Privileged observation is used as the ground truth for perception, while overall task success rate serves as the ground truth for planning and reasoning.

\textbf{Evaluation Metrics.} 
To assess the reliability of confidence estimates, we evaluate two key aspects: calibration and failure prediction~\cite{naeini2015obtaining, yuan2021large}. Calibration measures how well an agent's expressed confidence reflects its actual performance, \eg an 80\% confidence should ideally correspond to 80\% accuracy. This calibration is crucial for applications requiring robust risk assessment and trustworthiness. On the other hand, failure prediction focuses on the agent's ability to distinguish between correct and incorrect predictions by assigning higher confidence to correct outcomes.
We use the Expected Calibration Error (ECE) to quantify calibration quality and the Area Under the Receiver Operating Characteristic Curve (AUROC) to evaluate failure prediction. To address imbalances stemming from varying accuracy levels across tasks, we also include AUPRC-Positive (PR-P) and AUPRC-Negative (PR-N), which separately measure the agent's effectiveness in identifying correct and incorrect predictions.

\textbf{Minecraft Agents.} 
In this work, we focus on embodied agents powered by advanced Large Language Models (LLMs) and Vision-Language Models (VLMs) that enable multimodal reasoning and understanding in complex embodied environments. We employ three models as the agent's backbone: \textbf{(1) GPT-4V}, chosen for its strong performance in multimodal reasoning and proven effectiveness in complex environments like Minecraft ~\cite{wang2023voyager, qin2024mp5, li2024optimus} for planning and perception tasks. \textbf{(2) MineLLM} \cite{qin2024mp5}, a model specifically designed for Minecraft tasks, that leverages MineCLIP's visual encoder and Vicuna-13B \cite{vicuna2023} to deliver robust multimodal understanding. and \textbf{(3) STEVE}, built on the versatile LLaMA framework, STEVE models excel in contextual understanding and decision-making \cite{zhao2025see}. Fine-tuned for Minecraft, STEVE enhances planning, communication, and interaction capabilities. Detailed model descriptions are provided in Appendix \ref{appendix:model_description}.

\definecolor{bad}{RGB}{255, 210, 142}
\definecolor{bad_text}{RGB}{255, 220, 155}
\definecolor{borderline}{RGB}{255, 200, 170} 
\definecolor{good}{RGB}{255, 230, 180}

\begin{table*}[!ht]
\centering
\small
\begin{tabularx}{0.9\textwidth}{X X c c c c c}
\hline
\textbf{Metric} & \textbf{Model}    & \textbf{Vanilla} & \textbf{Self-Intervention} & \textbf{CoT (Inductive)} & \textbf{P\&S (Deductive)} & \textbf{Top-K (Abductive)} \\ \hline

\multirow{3}{*}{\textbf{ECE ↓}}  
                & GPT-4V           & 0.27             & 0.21               & 0.16          & \textbf{0.15}                       & 0.17           \\  
                & MineLLM          & 0.49             & 0.41               & \textbf{0.34}          & 0.39                       & 0.43           \\  
                & STEVE            & 0.43             & 0.32               & \textbf{0.26}          & \textbf{0.26}                       & 0.35           \\ \hline

\multirow{3}{*}{\textbf{AUROC ↑}} 
                & GPT-4V           & 0.69             & 0.76               & \textbf{0.83}          & 0.82                       & 0.73           \\  
                & MineLLM          & 0.53             & 0.59               & \textbf{0.64}          & 0.61                       & 0.58           \\  
                & STEVE            & 0.58             & 0.69               & \textbf{0.72}          & 0.67                       & 0.68           \\ \hline

\multirow{3}{*}{\textbf{PR-P ↑}}  
                & GPT-4V           & 0.66             & 0.76               & \textbf{0.81}          & 0.79                       & 0.70           \\  
                & MineLLM          & 0.51             & 0.59               & \textbf{0.63}          & 0.60                       & 0.57           \\  
                & STEVE            & 0.56             & 0.67               & \textbf{0.69}          & 0.66                       & 0.64           \\ \hline
                
\multirow{3}{*}{\textbf{PR-N ↑}}  
                & GPT-4V           & 0.52             & 0.53               & \textbf{0.58}          & 0.55                       & 0.53           \\  
                & MineLLM          & 0.39             & 0.42               & 0.42          & \textbf{0.43}                       & 0.40           \\  
                & STEVE            & 0.41             & 0.46               & \textbf{0.46}          & 0.43                       & 0.42           \\ \hline
\end{tabularx}
\vspace{-0.3cm}
\caption{\textbf{Confidence Metrics across Elicitation Policies} with three models (GPT-4V, MineLLM, and LLaMA-based STEVE) using different elicitation strategies: Vanilla (basic task understanding), Self-Intervention (reflection on own actions), Chain-of-Thought (step-by-step reasoning), Plan \& Solve (explicit planning before execution), and Top-K (confidence distribution across multiple outputs) with No Execution Policies applied. The best performance across each model is in \textbf{bold}.}
\label{tab:main_result}
\vspace{-0.3cm}
\end{table*}

\section{Experimental Results}
Table ~\ref{tab:main_result} presents the performance of benchmarked agents across all Elicitation Policies. In this experiment, evaluation is conducted without Execution Policies. The final confidence scores are computed as the average of individual step confidence scores across five independent task episodes.

\noindent \textbf{All Elicitation Policies Facilitate Better Calibration and Failure Prediction.} Across all models, Elicitation Policies consistently improve calibration (lower ECE) and failure prediction (higher AUROC, PR-P, PR-N) compared to the Vanilla baseline. For instance, in GPT-4V, every Elicitation Policy results in a lower ECE and higher AUROC relative to Vanilla, demonstrating their effectiveness in improving the robustness of uncertainty quantification. Likewise, MineLLM and STEVE exhibit noticeable gains in ECE and AUROC when incorporating elicitation mechanisms, confirming that Elicitation Policies help agents better assess uncertainty and predict incorrect responses.

\noindent \textbf{Structured Elicitation (CoT and P\&S) Improves Calibration and Failure Prediction the Most.} Among the four Elicitation Policies, structured reasoning approaches—CoT (Inductive) and P\&S (Deductive)—consistently yield the best calibration and failure detection performance. For example, in GPT-4V, P\&S achieves the lowest ECE (0.15) and one of the highest AUROC scores (0.82), while CoT further improves AUROC up to 0.83. Similar trends hold for MineLLM and STEVE, where CoT and P\&S outperform Self-Intervention and Top-K across nearly all metrics. These improvements suggest that breaking down reasoning into explicit steps helps the models maintain logical consistency, facilitating better overall calibration.

\noindent \textbf{Abductive Reasoning Poses Greater Challenges than Inductive and Deductive.} While Top-K (Abductive) improves over the Vanilla policy, it exhibits weaker calibration and failure prediction, suggesting that generating multiple plausible interpretations increases uncertainty misalignment, and therefore making it harder for the model to distinguish between correct and incorrect predictions. 
Additionally, the lower PR-P and PR-N scores indicate that confidence estimation for abductive reasoning is more difficult to calibrate compared to inductive and deductive settings.

\noindent \textbf{Confidence Calibration Remains Inconsistent Across Models.}  While GPT-4V consistently benefits from different Elicitation Policies, the improvements are less stable in fine-tuned models like MineLLM and STEVE. For instance, CoT boosts AUROC to 0.83 in GPT-4V but only reaches 0.64 in MineLLM and 0.72 in STEVE, indicating that fine-tuned models struggle to generalize confidence estimation effectively. One likely reason for this inconsistency is that MineLLM and STEVE, being fine-tuned models, exhibit degenerated language capabilities, limiting their ability to verbalize uncertainty reliably.

\definecolor{CustomRed}{RGB}{233, 93, 34}      
\definecolor{CustomOrange}{RGB}{245, 132, 38}  
\definecolor{CustomPeach}{RGB}{255, 165, 82}   
\definecolor{CustomLightOrange}{RGB}{255, 204, 128} 
\definecolor{CustomYellow}{RGB}{255, 220, 148} 
\definecolor{CustomDarkGray}{RGB}{80, 80, 80}
\definecolor{CustomGreen}{RGB}{0, 180, 0}

\begin{figure*}[!ht]
    \centering
    \begin{tikzpicture}
        \begin{axis}[
            hide axis,
            width=6cm, height=2cm, 
            xmin=0, xmax=1, ymin=0, ymax=1,
            legend style={at={(0.5,1.1)}, anchor=north, legend columns=-1, font=\color{CustomDarkGray}, draw=none}, 
        ]
            \addlegendimage{area legend, color=CustomRed, fill}
            \addlegendentry{Vanilla~~~}
            
            \addlegendimage{area legend, color=CustomOrange, fill}
            \addlegendentry{Self-Intervention~~~}
            
            \addlegendimage{area legend, color=CustomPeach, fill}
            \addlegendentry{CoT~~~}
            
            \addlegendimage{area legend, color=CustomLightOrange, fill}
            \addlegendentry{P\&S~~~}
            
            \addlegendimage{area legend, color=CustomYellow, fill}
            \addlegendentry{Top-K}
        \end{axis}
    \end{tikzpicture}

    
    \makebox[\textwidth][l]{
    \hspace{0.2cm}
    \begin{subfigure}{0.3\textwidth}
        \centering
        \begin{tikzpicture}
            \begin{axis}[
                ybar,
                ymin=0, ymax=1,
                xmin=1, xmax=2,
                width=7cm, height=3.5cm,
                bar width=6.05pt,  
                ylabel={ECE ↓},
                ylabel style={CustomDarkGray, font=\small},
                axis x line*=bottom, 
                axis y line*=left,
                axis line style={gray!60},
                xtick=\empty,
                xticklabels={Action Sampling, Scenario Reinterpretation, Hypothetical Reasoning},  
                xticklabel style={text width=1.6cm, align=center, font=\scriptsize, transparent},  
                yticklabel style={font=\small, gray},
                x=130pt,
                grid=major,
                major y grid style={gray!60, dashed, dash pattern=on 1pt off 1pt},
                major x grid style={transparent},
                nodes near coords,
                every node near coord/.append style={font=\scriptsize, rotate=90, anchor=west, gray,
                /pgf/number format/fixed,
                /pgf/number format/precision=2,
                /pgf/number format/fixed zerofill},
            ]
                \draw[dashed, red] (1,0.27) -- (2,0.27);
                \addplot[ybar, bar shift=-12pt] [fill=CustomRed, draw opacity=0] coordinates {(1.15,0.24) (1.5,0.22) (1.85,0.14)};
                \addplot[ybar, bar shift=-6pt] [fill=CustomOrange, draw opacity=0] coordinates {(1.15,0.10) (1.5,0.12) (1.85,0.11)};
                \addplot[ybar, bar shift=0pt] [fill=CustomPeach, draw opacity=0] coordinates {(1.15,0.11) (1.5,0.12) (1.85,0.11)};
                \addplot[ybar, bar shift=6pt] [fill=CustomLightOrange, draw opacity=0] coordinates {(1.15,0.11) (1.5,0.13) (1.85,0.18)};
                \addplot[ybar, bar shift=12pt] [fill=CustomYellow, draw opacity=0] coordinates {(1.15,0.12) (1.5,0.14) (1.85,0.15)};
        
            \end{axis}
        \end{tikzpicture}
        \vspace{-0.6cm}
    \end{subfigure}
    \begin{subfigure}{0.3\textwidth}
        \centering
        \begin{tikzpicture}
            \begin{axis}[
                ybar,
                ymin=0, ymax=1,
                xmin=1, xmax=2,
                width=7cm, height=3.5cm,
                bar width=6.05pt,  
                ylabel={ECE ↓},
                ylabel style={transparent},
                axis x line*=bottom, 
                axis y line*=left,
                axis line style={gray!60},
                xtick=\empty,
                xticklabels={Action Sampling, Scenario Reinterpretation, Hypothetical Reasoning},   
                xticklabel style={text width=1.6cm, align=center, font=\small, transparent},
                yticklabel style={font=\scriptsize, transparent},  
                x=130pt,
                grid=major,
                major y grid style={gray!60, dashed, dash pattern=on 1pt off 1pt},
                major x grid style={transparent},
                nodes near coords,
                every node near coord/.append style={font=\scriptsize, rotate=90, anchor=west, gray,
                /pgf/number format/fixed,
                /pgf/number format/precision=2,
                /pgf/number format/fixed zerofill},
            ]
                \draw[dashed, red] (1,0.49) -- (2,0.49);
                \addplot[ybar, bar shift=-12pt] [fill=CustomRed, draw opacity=0] coordinates {(1.15,0.50) (1.5,0.53) (1.85,0.52)};
                \addplot[ybar, bar shift=-6pt] [fill=CustomOrange, draw opacity=0] coordinates {(1.15,0.35) (1.5,0.42) (1.85,0.50)};
                \addplot[ybar, bar shift=0pt] [fill=CustomPeach, draw opacity=0] coordinates {(1.15,0.32) (1.5,0.44) (1.85,0.33)};
                \addplot[ybar, bar shift=6pt] [fill=CustomLightOrange, draw opacity=0] coordinates {(1.15,0.30) (1.5,0.45) (1.85,0.38)};
                \addplot[ybar, bar shift=12pt] [fill=CustomYellow, draw opacity=0] coordinates {(1.15,0.42) (1.5,0.47) (1.85,0.43)};
            \end{axis}
        \end{tikzpicture}
        \vspace{-0.6cm}
    \end{subfigure}
    \begin{subfigure}{0.3\textwidth}
        \centering
        \begin{tikzpicture}
            \begin{axis}[
                ybar,
                ymin=0, ymax=1,
                xmin=1, xmax=2,
                width=7cm, height=3.5cm,
                bar width=6.05pt,  
                ylabel={ECE ↓},
                ylabel style={transparent},
                axis x line*=bottom, 
                axis y line*=left,
                axis line style={gray!60},
                xtick=\empty,
                xticklabels={Action Sampling, Scenario Reinterpretation, Hypothetical Reasoning},   
                xticklabel style={text width=1.6cm, align=center, font=\small, transparent},  
                yticklabel style={font=\scriptsize, transparent},  
                x=130pt,
                grid=major,
                major y grid style={gray!60, dashed, dash pattern=on 1pt off 1pt},
                major x grid style={transparent},
                nodes near coords,
                every node near coord/.append style={font=\scriptsize, rotate=90, anchor=west, gray,
                /pgf/number format/fixed,
                /pgf/number format/precision=2,
                /pgf/number format/fixed zerofill},
            ]
                \draw[dashed, red] (1,0.43) -- (2,0.43);
                \addplot[ybar, bar shift=-12pt] [fill=CustomRed, draw opacity=0] coordinates {(1.15,0.40) (1.5,0.42) (1.85,0.41)};
                \addplot[ybar, bar shift=-6pt] [fill=CustomOrange, draw opacity=0] coordinates {(1.15,0.28) (1.5,0.27) (1.85,0.39)};
                \addplot[ybar, bar shift=0pt] [fill=CustomPeach, draw opacity=0] coordinates {(1.15,0.24) (1.5,0.24) (1.85,0.33)};
                \addplot[ybar, bar shift=6pt] [fill=CustomLightOrange, draw opacity=0] coordinates {(1.15,0.27) (1.5,0.28) (1.85,0.35)};
                \addplot[ybar, bar shift=12pt] [fill=CustomYellow, draw opacity=0] coordinates {(1.15,0.30) (1.5,0.31) (1.85,0.34)};
                
            \end{axis}
        \end{tikzpicture}
        \vspace{-0.6cm}
    \end{subfigure}
    }
    \makebox[\textwidth][l]{
    \hspace{0.2cm}
    \begin{subfigure}{0.3\textwidth}
        \centering
        \begin{tikzpicture}
            \begin{axis}[
                ybar,
                ymin=0, ymax=1.0,
                xmin=1, xmax=2,
                width=7cm, height=3.5cm,
                bar width=6.05pt,  
                ylabel={AUROC ↑},
                ylabel style={CustomDarkGray, font=\small},
                axis x line*=bottom, 
                axis y line*=left,
                axis line style={gray!60},
                xtick={1.15, 1.50, 1.85},  
                xtick style={draw=none},
                xticklabels={Action Sampling, Scenario Reinterpretation, Hypothetical Reasoning},   
                xticklabel style={text width=1.6cm, align=center, font=\scriptsize, gray},  
                yticklabel style={font=\small, gray},  
                x=130pt,
                grid=major,
                major y grid style={gray!60, dashed, dash pattern=on 1pt off 1pt},
                major x grid style={transparent},
                nodes near coords,
                every node near coord/.append style={font=\scriptsize, rotate=90, anchor=west, gray,
                /pgf/number format/fixed,
                /pgf/number format/precision=2,
                /pgf/number format/fixed zerofill},
            ]
                \draw[dashed, red] (1,0.69) -- (2,0.69);
                \addplot[ybar, bar shift=-12pt] [fill=CustomRed, draw opacity=0] coordinates {(1.15,0.76) (1.5,0.75) (1.85,0.78)};
                \addplot[ybar, bar shift=-6pt] [fill=CustomOrange, draw opacity=0] coordinates {(1.15,0.80) (1.5,0.79) (1.85,0.81)};
                \addplot[ybar, bar shift=0pt] [fill=CustomPeach, draw opacity=0] coordinates {(1.15,0.84) (1.5,0.82) (1.85,0.84)};
                \addplot[ybar, bar shift=6pt] [fill=CustomLightOrange, draw opacity=0] coordinates {(1.15,0.88) (1.5,0.79) (1.85,0.85)};
                \addplot[ybar, bar shift=12pt] [fill=CustomYellow, draw opacity=0] coordinates {(1.15,0.78) (1.5,0.77) (1.85,0.81)};
                
            \end{axis}
        \end{tikzpicture}
        \vspace{-0.5cm}
    \end{subfigure}
    \begin{subfigure}{0.3\textwidth}
        \centering
        \begin{tikzpicture}
            \begin{axis}[
                ybar,
                ymin=0, ymax=1.0,
                xmin=1, xmax=2,
                width=7cm, height=3.5cm,
                bar width=6.05pt,  
                ylabel={AUROC ↑},
                 ylabel style={transparent},
                axis x line*=bottom, 
                axis y line*=left,
                axis line style={gray!60},
                xtick={1.15, 1.50, 1.85},
                xtick style={draw=none},
                xticklabels={Action Sampling, Scenario Reinterpretation, Hypothetical Reasoning},   
                xticklabel style={text width=1.6cm, align=center, font=\scriptsize, gray},  
                yticklabel style={font=\small, transparent},  
                x=130pt,
                grid=major,
                major y grid style={gray!60, dashed, dash pattern=on 1pt off 1pt},
                major x grid style={transparent},
                nodes near coords,
                every node near coord/.append style={font=\scriptsize, rotate=90, anchor=west, gray,
                /pgf/number format/fixed,
                /pgf/number format/precision=2,
                /pgf/number format/fixed zerofill},
            ]
                
                \draw[dashed, red] (1,0.53) -- (2,0.53);
                \addplot[ybar, bar shift=-12pt] [fill=CustomRed, draw opacity=0] coordinates {(1.15,0.55) (1.5,0.64) (1.85,0.51)};
                \addplot[ybar, bar shift=-6pt] [fill=CustomOrange, draw opacity=0] coordinates {(1.15,0.64) (1.5,0.68) (1.85,0.68)};
                \addplot[ybar, bar shift=0pt] [fill=CustomPeach, draw opacity=0] coordinates {(1.15,0.69) (1.5,0.74) (1.85,0.71)};
                \addplot[ybar, bar shift=6pt] [fill=CustomLightOrange, draw opacity=0] coordinates {(1.15,0.67) (1.5,0.71) (1.85,0.65)};
                \addplot[ybar, bar shift=12pt] [fill=CustomYellow, draw opacity=0] coordinates {(1.15,0.63) (1.5,0.66) (1.85,0.65)};
                
            \end{axis}
        \end{tikzpicture}
        \vspace{-0.5cm}
    \end{subfigure}
    \begin{subfigure}{0.3\textwidth}
        \centering
        \begin{tikzpicture}
            \begin{axis}[
                ybar,
                ymin=0, ymax=1.0,
                xmin=1, xmax=2,
                width=7cm, height=3.5cm,
                bar width=6.05pt,  
                ylabel={AUROC ↑},
                ylabel style={transparent},
                axis x line*=bottom, 
                axis y line*=left,
                axis line style={gray!60},
                xtick={1.15, 1.50, 1.85},  
                xtick style={draw=none},
                xticklabels={Action Sampling, Scenario Reinterpretation, Hypothetical Reasoning},   
                xticklabel style={text width=1.6cm, align=center, font=\scriptsize, gray},  
                yticklabel style={font=\small, transparent},  
                x=130pt,
                grid=major,
                major y grid style={gray!60, dashed, dash pattern=on 1pt off 1pt},
                major x grid style={transparent},
                nodes near coords,
                every node near coord/.append style={font=\scriptsize, rotate=90, anchor=west, gray,
                /pgf/number format/fixed,
                /pgf/number format/precision=2,
                /pgf/number format/fixed zerofill},
            ]
                \draw[dashed, red] (1,0.58) -- (2,0.58);
                \addplot[ybar, bar shift=-12pt] [fill=CustomRed, draw opacity=0] coordinates {(1.15,0.62) (1.5,0.66) (1.85,0.57)};
                \addplot[ybar, bar shift=-6pt] [fill=CustomOrange, draw opacity=0] coordinates {(1.15,0.74) (1.5,0.74) (1.85,0.65)};
                \addplot[ybar, bar shift=0pt] [fill=CustomPeach, draw opacity=0] coordinates {(1.15,0.72) (1.5,0.70) (1.85,0.65)};
                \addplot[ybar, bar shift=6pt] [fill=CustomLightOrange, draw opacity=0] coordinates {(1.15,0.72) (1.5,0.72) (1.85,0.64)};
                \addplot[ybar, bar shift=12pt] [fill=CustomYellow, draw opacity=0] coordinates {(1.15,0.70) (1.5,0.68) (1.85,0.64)};
                
            \end{axis}
        \end{tikzpicture}
        \vspace{-0.5cm}
    \end{subfigure}
    }
    \makebox[\textwidth][r]{
        \begin{minipage}{0.33\textwidth}
            \centering \textcolor{CustomDarkGray}{GPT-4V}
        \end{minipage}
        \hspace{-0.5cm}
        \begin{minipage}{0.33\textwidth}
            \centering \textcolor{CustomDarkGray}{MineLLM}
        \end{minipage}
        \hspace{-0.5cm}
        \begin{minipage}{0.33\textwidth}
            \centering \textcolor{CustomDarkGray}{STEVE}
        \end{minipage}
        \hspace{-0.3cm}
    }
    \vspace{-0.7cm}
    \caption{\textbf{ECE and AUROC across Models and Execution Policies.} Bars present ECE (top, lower is better) and AUROC (bottom, higher is better) under different elicitation strategies. \textbf{\textcolor{red}{Red dashed lines}} are metrics for Vanilla elicitation with no execution policy applied.}
    \label{fig:main_bar}
    \vspace{-0.3cm}

\end{figure*}
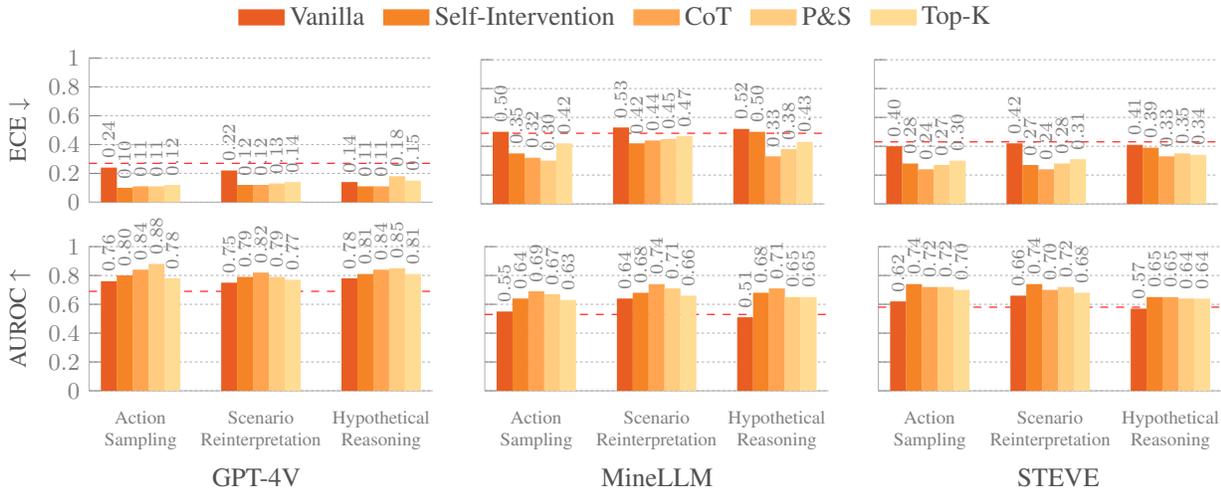

\noindent \textbf{Execution Policies Amplify Reliable Embodied Confidence Across Elicitation Policies.} Figure \ref{fig:main_bar} illustrates how Execution Policies interact with Elicitation Policies. Overall, Execution Policies are capable of further improving calibration and failure prediction performance. For example, GPT-4V achieves better ECE results when pairing any Execution Policy with all Elicitation Policies.
More specifically, structured reasoning approaches such as CoT (Inductive) and P\&S (Deductive), when paired with Action Sampling, tend to yield improved confidence calibration. For instance, MineLLM's ECE achieves 0.32 and 0.30 paired with CoT and P\&S respectively, outperforming other combinations. Hypothetical Reasoning sometimes degrades performance. For instance, STEVE's ECE worsens when pairing Hypothetical Reasoning with all Elicitation Policies, suggesting that while this execution strategy allows models to reason over multiple possible outcomes, it may introduce uncertainty, leading to less calibrated confidence judgments.

\textbf{So, How Effectively Can Embodied Agents Express Confidence in Dynamic Embodied Tasks?} While embodied agents can convey confidence to some extent, their effectiveness depends on how well they integrate reasoning, uncertainty assessment, and environmental interactions. The findings reveal that embodied confidence elicitation remains a challenging problem, requiring a careful balance between general-purpose reasoning and task-specific specialization. However, our proposed Elicitation Policies improve both confidence calibration and failure prediction, while our Execution Policies further augment these performance gains by refining uncertainty through iterative interactions with the environment. These results highlight the importance of accounting for the unique challenges faced by embodied agents in confidence estimation, emphasizing the need for execution-aware strategies that enhance both calibration and failure prediction in complex environments.

\section{Ablation Studies}
\textbf{Impact of Execution Policies.}
We analyze the performance of Execution Policy combinations, incorporating Action Sampling (AS), Scenario Reinterpretation (SR), and Hypothetical Reasoning (HR) both incrementally and collectively. Results in Table \ref{tab:execu_ablation} show clear trends in how Execution Policies influence performance. Without any Execution Policies, Vanilla Elicitation exhibits the worst calibration, with ECE as high as 0.27, while also struggling with failure prediction. When Execution Policies are introduced, performance improves, though trade-offs emerge between failure prediction accuracy (AUROC) and confidence calibration (ECE). Among two-policy combinations, the combination of Action Sampling with Scenario Reinterpretation (AS + SR) delivers the most balanced improvement, significantly increasing AUROC (up to 0.83 for GPT-4V and 0.69 for STEVE) while maintaining the lowest ECE (0.17 for GPT-4V, 0.32 for MineLLM). This suggests that jointly exploring multiple action paths and reinterpreting environmental cues helps refine confidence estimation without sacrificing calibration. 

In addition, strategies incorporating Action Sampling (AS) consistently outperform those without it, resulting in better uncertainty estimation and more reliable confidence scores. By generating multiple action plans, AS enhances confidence calibration, underscoring the importance of addressing action planning uncertainty in embodied agents.
Combining all Execution Policies yields the strongest overall performance across models, achieving the highest AUROC across all three models while maintaining competitive calibration, with the lowest ECE for GPT-4V (0.17) and strong values for MineLLM (0.32) and STEVE (0.38).  This suggests that integrating Action Sampling, Scenario Reinterpretation, and Hypothetical Reasoning provides a complementary effect, improving both failure prediction accuracy and confidence estimation.
 
\begin{table*}[!ht]
    \centering 
    \begin{tabularx}{0.9\textwidth}{l X X X X X X}
        \hline
        \multirow{2}{*}{\textbf{Execution Strategies}}
         & \multicolumn{2}{c}{\textbf{GPT-4V}} & \multicolumn{2}{c}{\textbf{MineLLM}} & \multicolumn{2}{c}{\textbf{STEVE}} \\ 
        \cmidrule(lr){2-3} \cmidrule(lr){4-5} \cmidrule(lr){6-7}
         & \textbf{ECE ↓} & \textbf{AUROC ↑} & \textbf{ECE ↓} & \textbf{AUROC ↑} & \textbf{ECE ↓} & \textbf{AUROC ↑} \\ 
        \hline
        No Execution Strategy  & 0.27 & 0.69 & 0.49 & 0.53 & 0.43 & 0.58 \\ 
        AS + SR  & 0.18 & 0.82 & \textbf{0.32} & 0.59 & 0.39 & 0.69 \\ 
        AS + HR  & 0.20 & 0.79 & 0.34 & 0.57 & \textbf{0.37} & 0.66 \\ 
        SR + HR  & 0.22 & 0.80 & 0.37 & 0.54 & 0.44 & 0.58 \\ 
        AS + SR + HR  & \textbf{0.17} & \textbf{0.83} & \textbf{0.32} & \textbf{0.62} & 0.38 & \textbf{0.69} \\ 
        \hline
    \end{tabularx}
    \vspace{-0.3cm}
    \caption{\textbf{Performance of Vanilla Elicitation with Combined Execution Strategies.}
    \textbf{AS} = Action Sampling, \textbf{SR} = Scenario Reinterpretation, \textbf{HR} = Hypothetical Reasoning. 
    ECE and AUROC for each model, GPT-4V, MineLLM, and STEVE. 
    Best values highlighted in \textbf{bold}.}
    \label{tab:execu_ablation}
    \vspace{-0.3cm}
\end{table*}

\textbf{Perception \vs Cognition.}
Embodied agents, when tasked with high-level objectives (\eg ``find a pig"), often rely on language models to decompose the task into smaller, granular actions (\eg ``step forward 2 steps"). During task execution, the agent generates confidence scores for each granular action. Typically, these scores are aggregated temporally to produce a single overall confidence score for the entire task. While this method provides a holistic measure of confidence, it does not differentiate between the confidence associated with perception (\eg recognizing a pig) and cognition (\eg reasoning about the sequence of steps).

\begin{figure}[tp!]
    \centering
    \begin{subfigure}[b]{0.8\columnwidth}
        \centering
        \begin{tikzpicture}
            \begin{axis}[
                hide axis,
                width=8.5cm, height=2cm, 
                xmin=0, xmax=1, ymin=0, ymax=1,
                legend style={at={(0.5,1.1)}, anchor=north, legend columns=-1, font=\scriptsize \color{CustomDarkGray}, draw=none}, 
            ]
                \addlegendimage{area legend, color=CustomRed, fill}
                \addlegendentry{ECE~~~}
                
                \addlegendimage{area legend, color=CustomOrange, fill}
                \addlegendentry{AUROC~~~}
                
                \addlegendimage{area legend, color=CustomPeach, fill}
                \addlegendentry{PR-P~~~}
                
                \addlegendimage{area legend, color=CustomLightOrange, fill}
                \addlegendentry{PR-N}
            \end{axis}
        \end{tikzpicture}
    \end{subfigure}

    \begin{subfigure}[b]{\columnwidth}
        \begin{tikzpicture}
            \begin{axis}[
                ybar,
                ymin=0, ymax=1.0,
                xmin=1, xmax=2,
                 height=3.5cm,
                bar width=8pt,  
                ylabel={Confidence Scores},
                ylabel style={CustomDarkGray, font=\small},
                axis x line*=bottom, 
                axis y line*=left,
                axis line style={gray!60},
                xtick={1.15, 1.50, 1.85},
                xtick style={draw=none},
                xticklabels={Temporal Aggregation, Reasoning Confidence, Perception Confidence},   
                xticklabel style={text width=1.6cm, align=center, font=\small, CustomDarkGray},  
                yticklabel style={font=\small, gray},  
                x=180pt,
                grid=major,
                major y grid style={gray!60, dashed, dash pattern=on 1pt off 1pt},
                major x grid style={transparent},
                nodes near coords,
                every node near coord/.append style={font=\small, rotate=90, anchor=west, gray, 
                /pgf/number format/fixed,
                /pgf/number format/precision=2,
                /pgf/number format/fixed zerofill},
            ]
                \addplot[ybar, bar shift=-11.8pt] [fill=CustomRed, draw opacity=0] coordinates {(1.15,0.18) (1.5,0.26) (1.85,0.22)};
                \addplot[ybar, bar shift=-3.9pt] [fill=CustomOrange, draw opacity=0] coordinates {(1.15,0.76) (1.5,0.71) (1.85,0.79)};
                \addplot[ybar, bar shift=3.9pt] [fill=CustomPeach, draw opacity=0] coordinates {(1.15,0.82) (1.5,0.78) (1.85,0.85)};
                \addplot[ybar, bar shift=11.8pt] [fill=CustomLightOrange, draw opacity=0] coordinates {(1.15,0.80) (1.5,0.72) (1.85,0.81)};   
            \end{axis}
        \end{tikzpicture}
    \end{subfigure}
    \vspace{-0.9cm}
    \caption{\textbf{Comparison of Aggregation Methods for Vanilla Elicitation without Execution Policies.} Temporal aggregation provides a holistic score, while separate aggregation evaluates confidence in reasoning and perception separately respectively.}
    \label{fig:aggregation_methods}
    \vspace{-0.5cm}
\end{figure}
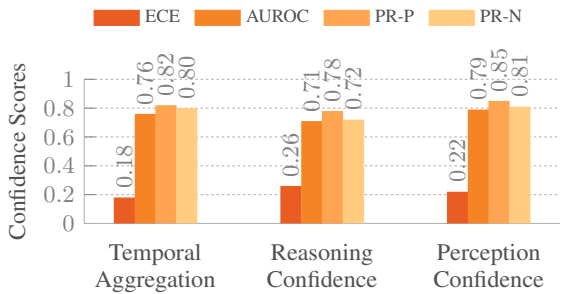

To better understand how different sources of uncertainty contribute to overall confidence, we separately analyze perception and reasoning confidence. Perception Confidence aggregates scores related to the agent's ability to interpret its sensory inputs (\eg detecting objects or understanding environmental cues), while {Reasoning Confidence} aggregates scores associated with reasoning and decision-making processes during task execution.
Figure \ref{fig:aggregation_methods} reveals that temporal aggregation achieves the lowest ECE (0.18) and a balanced AUROC (0.76). Temporal aggregation smooths over individual uncertainties, providing robust overall calibration and reliable failure prediction.

Perception-based confidence, when aggregated separately, offers a distinct advantage in predictive reliability. With an AUROC of 0.79, the highest among the methods, and strong PR-P (0.85) and PR-N (0.81) scores, perception confidence consistently outperforms reasoning. This highlights the inherent stability of sensory tasks, where clear input-output mappings and deterministic operations reduce uncertainty. Additionally, perception confidence maintains a competitive ECE (0.22), indicating that it remains well-calibrated. 

In contrast, reasoning confidence introduces more uncertainty, resulting in a higher ECE (0.26) and a lower AUROC (0.71). These results reflect the challenges of reasoning tasks, which often involve multi-step decision-making and are susceptible to cascading errors. The lower PR-P (0.78) and PR-N (0.72) scores suggest reasoning confidence struggles to accurately distinguish correct from incorrect outcomes. In essence, results affirm that reasoning tasks inherently present greater uncertainty, requiring more sophisticated calibration methods to maintain reliability.

Interestingly, the performance gap between perception and reasoning confidence underscores their complementary nature. While perception excels in calibration and failure prediction, reasoning provides critical insights into decision-making under uncertainty. Temporal aggregation balances these components effectively for an overall confidence score but sacrifices the interpretability offered by separate aggregation. This comparison emphasizes the need to align aggregation methods with task complexity and performance priorities, whether for holistic confidence measures or detailed insights into perception and cognition.

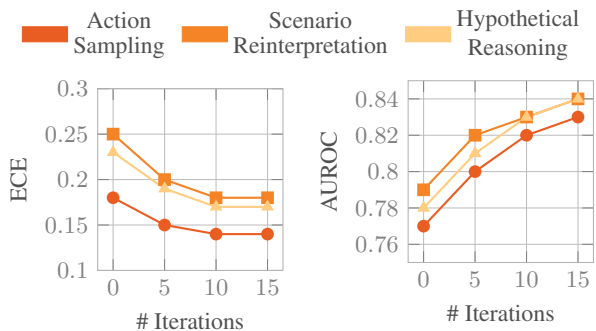
\begin{figure}
    \centering
    \begin{tikzpicture}
        \begin{axis}[
            hide axis,
            width=\columnwidth,
            height=2cm, 
            xmin=0, xmax=1, ymin=0, ymax=1,
            legend style={at={(0.5,1.1)}, anchor=north, legend columns=-1, font=\color{CustomDarkGray}, draw=none, font=\small, text=CustomDarkGray}, 
        ]
            \addlegendimage{area legend, color=CustomRed, fill}
            \addlegendentry{\shortstack{Action\\Sampling}~~~}
            
            \addlegendimage{area legend, color=CustomOrange, fill}
            \addlegendentry{\shortstack{Scenario\\Reinterpretation}~~~}
            
            \addlegendimage{area legend, color=CustomLightOrange, fill}
            \addlegendentry{\shortstack{Hypothetical\\Reasoning}}   
        \end{axis}
    \end{tikzpicture}
    
    \begin{minipage}{0.49\columnwidth}
        \centering
        \begin{tikzpicture}
            \begin{axis}[
                width=\columnwidth,
                height=4cm,
                axis line style={gray!60},
                xlabel={\# Iterations},
                ylabel={ECE},
                label style = {CustomDarkGray},
                xtick=data,
                ymin=0.1, ymax=0.3,
                grid=major,
                label style={font=\small},
                tick label style={font=\small, gray},
            ]
            \addplot[thick, mark=*, color=CustomRed] coordinates {(0,0.18) (5,0.15) (10,0.14) (15,0.14)};
            \addplot[thick, mark=square*, color=CustomOrange] coordinates {(0,0.25) (5,0.20) (10,0.18) (15,0.18)};
            \addplot[thick, mark=triangle*, color=CustomLightOrange] coordinates {(0,0.23) (5,0.19) (10,0.17) (15,0.17)};
            \end{axis}
        \end{tikzpicture}
    \end{minipage}
    \begin{minipage}{0.49\columnwidth}
        \centering
        \begin{tikzpicture}
            \begin{axis}[
                width=\columnwidth,
                height=4cm,
                axis line style={gray!60},
                xlabel={\# Iterations},
                ylabel={AUROC},
                label style = {CustomDarkGray},
                xtick=data,
                ymin=0.75, ymax=0.85,
                grid=major,
                label style={font=\small},
                tick label style={font=\small, gray}
            ]
            \addplot[thick, mark=*, color=CustomRed] coordinates {(0,0.77) (5,0.80) (10,0.82) (15,0.83)};
            \addplot[thick, mark=square*, color=CustomOrange] coordinates {(0,0.79) (5,0.82) (10,0.83) (15,0.84)};
            \addplot[thick, mark=triangle*, color=CustomLightOrange]  coordinates {(0,0.78) (5,0.81) (10,0.83) (15,0.84)};
            \end{axis}
        \end{tikzpicture}
    \end{minipage}
    \vspace{-0.3cm}
    \caption{\textbf{ECE and AUROC across Execution Policy Iterations.}}
    \label{fig:line_charts}
    \vspace{-0.5cm}
\end{figure}
\textbf{Impact of Execution Iterations.} We investigate the impact of repeatedly applying execution policies on calibration and failure prediction accuracy. Iterations range from 0 (\ie no execution policies employed) to 15, allowing for an analysis of both the initial benefits and potential diminishing returns of repeated applications.
As shown in Figure \ref{fig:line_charts}, repeated applications initially improve ECE across all policies but eventually plateau. For instance, Action Sampling reduces ECE from 0.18 (at 0 iterations) to 0.14 (at 15 iterations), with most of the improvement occurring within the first 10 iterations. A similar trend is observed for Scenario Reinterpretation and Hypothetical Reasoning, where ECE drops from 0.25 to 0.18 and from 0.23 to 0.17, respectively. The plateau effect is less pronounced in AUROC, which consistently improves across iterations. Action Sampling increases AUROC from 0.77 to 0.83, while Scenario Reinterpretation and Hypothetical Reasoning improve from 0.79 to 0.84 and from 0.78 to 0.84, respectively. Most AUROC gains occur between 0 and 10 iterations, with diminishing returns after 15 iterations.
Overall, early iterations improve calibration and failure prediction, but excessive repetition yields diminishing returns. This underscores the need to balance execution policy applications for optimal effectiveness.

\section{Conclusion}
This work presents the first systematic exploration of embodied confidence elicitation, introducing elicitation and execution policies that enhance calibration and failure prediction in open-ended multimodal embodied tasks. Our findings highlight improvements in confidence estimation using our proposed methods, providing more accurate uncertainty quantification. Future research could improve confidence elicitation in embodied environments by scaling to more diverse and complex environments and exploring their integration with various embodied agent architectures. 

\section{Impact Statement}
This work advances Embodied AI by introducing confidence elicitation and execution policies tailored to multimodal and dynamic environments. By enabling embodied agents to express uncertainty, our approach enhances their calibration, adaptability, and reliability in complex tasks. This contribution supports safer AI deployment in real-world domains like robotics, education, and collaborative systems, where accurate self-assessment is critical. However, the reliance on large pre-trained models raises concerns about energy efficiency and ethical considerations in high-stakes applications, which warrant further exploration.

\bibliography{reference}
\bibliographystyle{icml2025}

\newpage
\appendix
\onecolumn

\section{Definitions of Uncertainty Types}
\label{appendix:uncertainty_definition} The three fundamental forms of logical reasoning, inductive, deductive, and abductive, have long been recognized and studied~\cite{peirce1934collected, walton2014abductive, wei2022chain, levine2022inductivebiasincontextlearning, okoli2023inductive}. As language models demonstrated extraordinary abilities, designing better reasoning mechanisms has become a popular research trend~\cite{cheng2024inductive, liu2024incomplete}. These reasoning paradigms serve as fundamental frameworks for structuring inference and decision-making processes, particularly in settings where uncertainty arises due to partial observations, ambiguous premises, or multiple plausible explanations~\cite{xu2025large}. We adapt these reasoning types to the domain of embodied confidence elicitation and formally define and describe each uncertainty type (See Table~\ref{tab:uncertainty_definition}). 

\begin{table*}[!ht]
    \centering
    \renewcommand{\arraystretch}{1.5} 
    \small
    \begin{tabularx}{\textwidth}{p{1cm} X X X X X}
        \multicolumn{1}{c}{\textbf{Reasoning Type}} & 
        \multicolumn{1}{c}{\textbf{Definition}} & 
        \multicolumn{1}{c}{\textbf{Uncertainty Associated}} & 
        \multicolumn{1}{c}{\textbf{Example}} & 
        \multicolumn{1}{c}{\textbf{Elicitation Method}} \\
        \hline
        \multirow{7}{*}{\textbf{Inductive}} & 
         Inductive reasoning derives general principles from a body of observations which means making broad generalizations based on specific observations ~\cite{Li1992InductiveR}. & 
        \textit{Inductive Uncertainty}: Arises when an agent generalizes from limited observations, leading to potential overgeneralization or misclassification. & 
        An agent observes that all previously encountered caves contained hostile entities and infers that any future cave is also dangerous. However, this conclusion is uncertain because it is based on a limited number of observations. & 
        \textit{Chain-of-Thought (CoT)}: The agent systematically analyzes observed trends, considers possible exceptions, and evaluates confidence in applying generalizations. \\
        \hline
        \multirow{7}{*}{\textbf{Deductive}} & 
        Deductive reasoning is the process of drawing deductive inferences that start from the given premises and reason with logical rules or commonsense to obtain certain conclusions ~\cite{johnson1999deductive, goel2007anatomy}. & 
        \textit{Deductive Uncertainty}: Arises when an agent applies logical rules but encounters missing, conflicting, or incomplete premises, making the outcome uncertain. & 
        An agent knows the rule: "If wood is available, then a wooden tool can be crafted." However, if it is uncertain whether wood is available, it cannot confidently conclude whether crafting is possible. & 
        \textit{Plan-and-Solve (P\&S)}: The agent formulates a set of premises, identifies missing dependencies, and assesses confidence in executing the task. \\
        \hline
        \multirow{7}{*}{\textbf{Abductive}} & 
        The process of inferring the most plausible explanation for an observation based on incomplete evidence. Abduction generates hypotheses rather than definitive conclusions ~\cite{josephson1996abductive, walton2001abductive}. & 
        \textit{Abductive Uncertainty}: Arises when multiple explanations could account for an observation, with no definitive way to determine the correct one. & 
        An agent searching for a pig near a river hypothesizes that pigs and rivers may exist in any of the four cardinal directions but lacks direct evidence to confirm a single hypothesis. & 
        \textit{Top-K Reasoning}: The agent generates multiple plausible hypotheses, assigns probability estimates to each, and ranks them by likelihood. \\
        \hline
    \end{tabularx}
    \caption{Definitions of reasoning types, their associated uncertainty, examples, and the corresponding elicitation methods.}
    \label{tab:uncertainty_definition}
\end{table*}

\noindent \textbf{Inductive Uncertainty} arises when an agent generalizes from specific observations to broader conclusions based on incomplete data. Induction relies on identifying patterns from limited experiences, leading to inherent uncertainty. This is particularly relevant in open-world environments, where observations are partial, and inferred generalizations may not always hold. For example, an agent navigating an unfamiliar environment may observe that all previously encountered caves contained hostile entities. Based on this pattern, it may infer that any future cave is also dangerous. However, since this conclusion is based on a limited set of observations rather than a deterministic rule, the agent must assess how strongly its past experiences justify this generalization, introducing \textit{inductive uncertainty}.

To elicit inductive uncertainty, we employ Chain-of-Thought \cite{wei2022chain}, which prompts the agent to explicitly reflect on the reliability of its observed patterns. By systematically verbalizing its reasoning, the agent is encouraged to: (1) analyze the strength of observed trends, (2) consider possible exceptions or contradictory evidence, and (3) assess its confidence in applying the generalization to new situations. This structured elicitation enables the agent to express uncertainty in its inductive inferences rather than assuming patterns always hold.

\noindent \textbf{Deductive Uncertainty} arises when an agent faces ambiguity due to missing, conflicting, or incomplete premises. Deductive uncertainty occurs within a structured decision-making process when the available information is insufficient to determine a definitive outcome. Consider an agent tasked with crafting a wooden tool in a survival environment. It knows the rule: "If wood is available, then a wooden tool can be crafted." However, if the agent is uncertain whether wood is currently accessible, it cannot confidently conclude whether crafting is possible. This scenario exemplifies deductive uncertainty, where the agent’s ability to reason is constrained by unknown or ambiguous premises.

To elicit deductive uncertainty, we use \textit{Plan-and-Solve} prompting \cite{wang2023plan}, which guides the agent through a structured reasoning process. The agent is encouraged to: (1) formulate a comprehensive set of premises relevant to the task, (2) identify any missing premises or dependencies, and (3) assess its confidence in executing each step successfully. This structured elicitation enables the agent to explicitly express uncertainty when premises are incomplete or insufficient to deduce a definitive conclusion.

\noindent \textbf{Abductive Uncertainty} occurs when an agent must infer the most plausible explanation for an observation without definitive evidence. Abduction involves \textit{hypothesis generation} under uncertainty. The challenge in abductive reasoning lies in selecting the most probable explanation when multiple interpretations exist, each carrying some degree of uncertainty. A simple example occurs when an agent is tasked with locating a pig near a river for unspecified reasons. Given its environment, the agent may hypothesize that pigs and rivers could exist in any of the four cardinal directions but are unlikely to be present in all directions simultaneously. Since the agent lacks direct evidence to confirm a single hypothesis, it must infer the most plausible explanation, leading to abductive uncertainty.

To elicit abductive uncertainty, we implement \textit{Top-K reasoning} \cite{robinson2022leveraging}, where the agent is instructed to generate multiple plausible hypotheses explaining an observation and assign probability estimates to each. This process forces the agent to explicitly consider alternative interpretations, rank them by likelihood, and communicate the level of confidence in its inferences. By quantifying uncertainty across multiple competing hypotheses, Top-K reasoning reveals the agent’s abductive reasoning capabilities.

\section{Task Setting Details} 
\label{appendix:task_setting}
Inspired by previous works \cite{lin2023mcu, qin2024mp5}, we define a set of 30 tasks evenly distributed across three difficulty levels: easy, medium, and hard. Categorization is based on the complexity of reasoning required and the extent of contextual information necessary for successful task completion. Each difficulty level incorporates distinct challenges, ranging from straightforward operations to intricate reasoning across interdependent objectives, with a balanced distribution of complexity within the task set. We present all tasks and highlight entities of different categories in Table \ref{tab:task_settings}

\noindent \textbf{Easy Tasks} are designed to evaluate the agent’s ability to process minimal perceptual information and perform straightforward actions with limited reasoning. These tasks typically require the perception of only one environmental element from predefined categories such as Object, Mob, Ecology, Time, Weather, or Brightness \cite{qin2024mp5}. For example, tasks at this level may involve identifying a specific object in the environment or recognizing a simple temporal condition (\eg daytime or nighttime). The actions required are relatively simple and involve a single reasoning step, such as gathering an object that is readily visible.

\noindent \textbf{Medium Tasks} introduce moderate complexity by requiring the perception and integration of two to three environmental elements, alongside a corresponding increase in reasoning steps. Tasks at this level involve combining multiple types of perceptual data, such as recognizing a specific biome and locating a particular mob or object within it. For example, the agent might need to identify a forest biome, locate a pig, and gather specific materials. In addition to perceptual challenges, medium tasks often include sequential sub-goals, such as collecting and combining resources to create basic tools. These tasks require the agent to interpret dynamic environmental information, execute plans involving multiple steps, and adapt to minor changes in the environment. This level evaluates the agent’s ability to balance perception, reasoning, and adaptability.

\noindent \textbf{Hard Tasks} are the most challenging and require the agent to process and integrate multiple layers of perceptual information (up to six elements) while performing complex situation-aware planning and dynamic action execution. These tasks involve a high level of reasoning, such as decomposing long-term objectives into interdependent sub-tasks, managing uncertainties in the environment, and dynamically adjusting strategies in response to real-time changes. For example, a hard task might involve navigating through hazardous biomes, identifying and gathering multiple resources, and crafting advanced tools or items that require sequential processing and the use of specialized platforms. Environmental conditions, such as weather, time of day, or changing brightness, may dynamically impact the task, necessitating constant adaptation by the agent. These tasks often introduce significant challenges, such as hostile mobs or the need to traverse difficult terrain, testing the agent's ability to balance perception, planning, and execution effectively.

\begin{figure}[t!]
    \centering
    \includegraphics[width=0.8\textwidth]{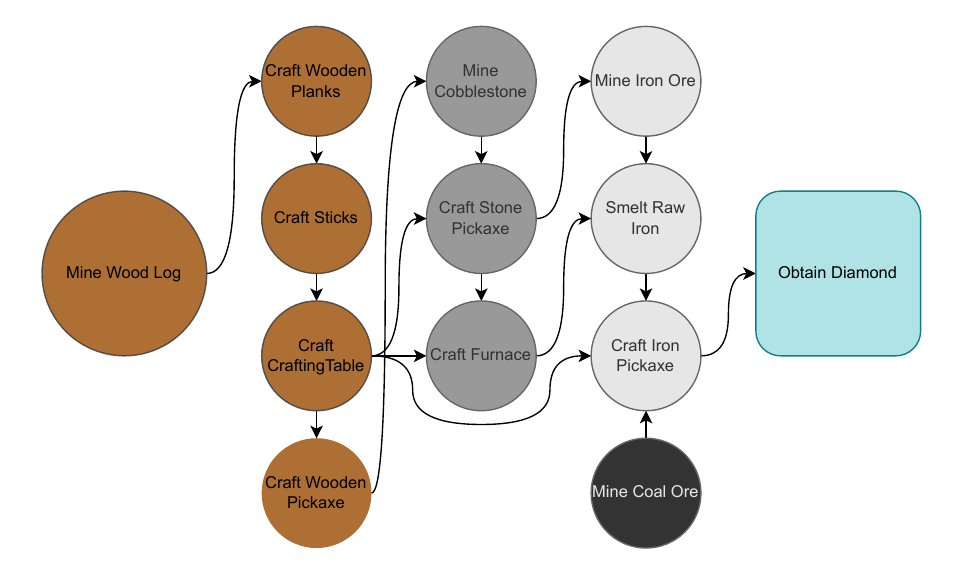} 
    \caption{An illustrative diagram of the Obtain Diamond task, featuring five distinct colors to represent the source materials required—wood, stone, iron, coal, and diamond—aligned with the Minecraft tech tree.}
    \label{fig:diamond_challenge} 
\end{figure}

\noindent The (obtain) \textbf{Diamond Task} is one of the most iconic and challenging benchmarks in Minecraft agent research, serving as a comprehensive test of an agent's long-horizon planning, resource management, and adaptability. The task requires the agent to progress through multiple interdependent steps, including gathering basic resources like wood and stone, crafting tools such as a pickaxe, and locating and mining diamonds deep within underground caves (See Figure \ref{fig:diamond_challenge}). Each step presents its own set of challenges, such as navigating complex terrain, managing limited resources, and avoiding environmental hazards like lava or hostile mobs. The randomized nature of Minecraft's procedural world generation further compounds the difficulty, as the agent must adapt dynamically to new environments while maintaining focus on the ultimate objective. Success in the ``Obtain Diamond" task is often seen as a key indicator of an agent's ability to integrate active perception, situational awareness, and embodied action execution in an open-world setting. This task demonstrates the complexity of open-ended problem-solving and has become a gold standard for evaluating the capabilities of autonomous agents in multi-modal and multi-step scenarios. We added the diamond task as one of our hard tasks.

\begin{table}[htbp]
    \centering
    \renewcommand{\arraystretch}{1.5} 
    \setlength{\tabcolsep}{5pt} 
    \begin{tabular}{c c p{10cm}<{\centering}}
        
        \textbf{Difficulty} & \textbf{Task ID} & \textbf{Task Description} \\ \hline
        \multirow{10}{*}{Easy} 
        & 1  & Find a \underline{pig} \raisebox{-0.3\height}{\includegraphics[width=0.5cm]{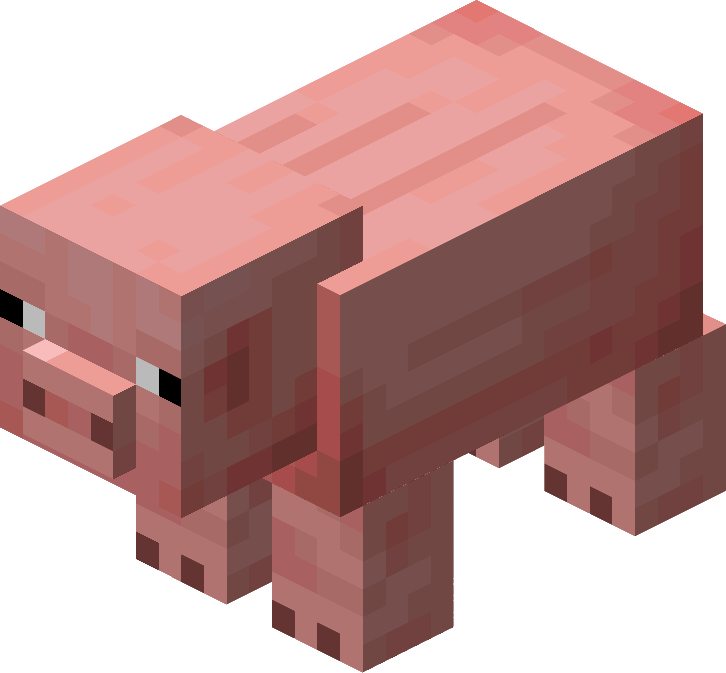}} \\ 
        & 2  & Find a \underline{cow} \raisebox{-0.3\height}{\includegraphics[width=0.5cm]{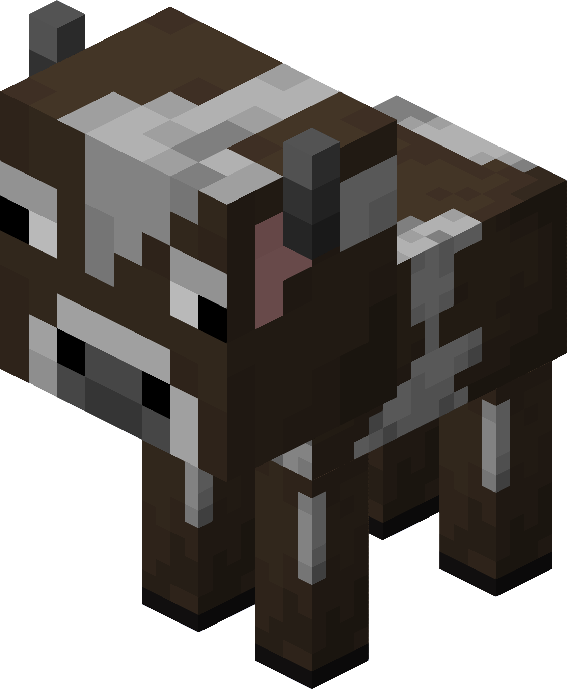}} \\ 
        & 3  & Find a \underline{tree} \raisebox{-0.3\height}{\includegraphics[width=0.5cm]{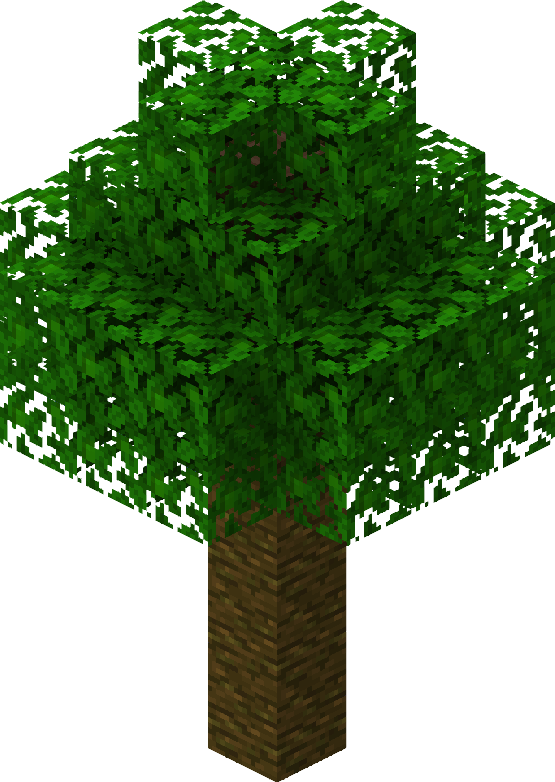}} \\
        & 4  & Mine \underline{log} \raisebox{-0.3\height}{\includegraphics[width=0.5cm]{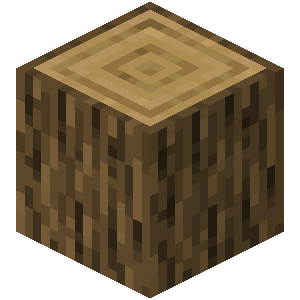}} \\ 
        & 5  & Mine \underline{sand} \raisebox{-0.3\height}{\includegraphics[width=0.5cm]{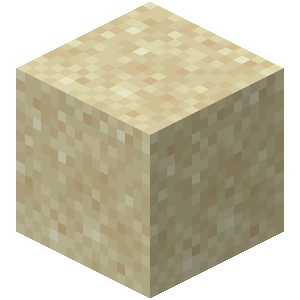}} \\ 
        & 6  & Craft a \underline{plank} \raisebox{-0.3\height}{\includegraphics[width=0.5cm]{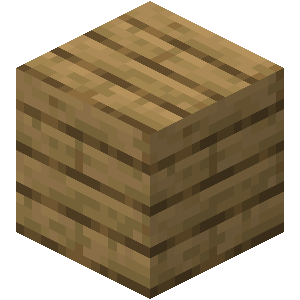}} \\ 
        & 7  & Craft a \underline{stick} \raisebox{-0.3\height}{\includegraphics[width=0.5cm]{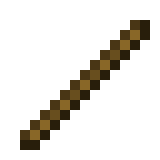}} \\ 
        & 8  & Craft a \underline{chest} \raisebox{-0.3\height}{\includegraphics[width=0.5cm]{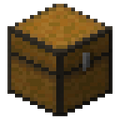}} \\ 
        & 9  & Craft a \underline{wooden door} \raisebox{-0.3\height}{\includegraphics[width=0.5cm]{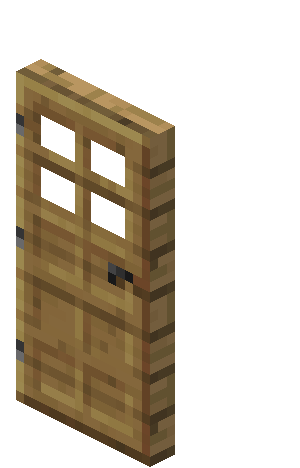}} \\ 
        & 10 & Craft a \underline{wooden boat} \raisebox{-0.3\height}{\includegraphics[width=0.5cm]{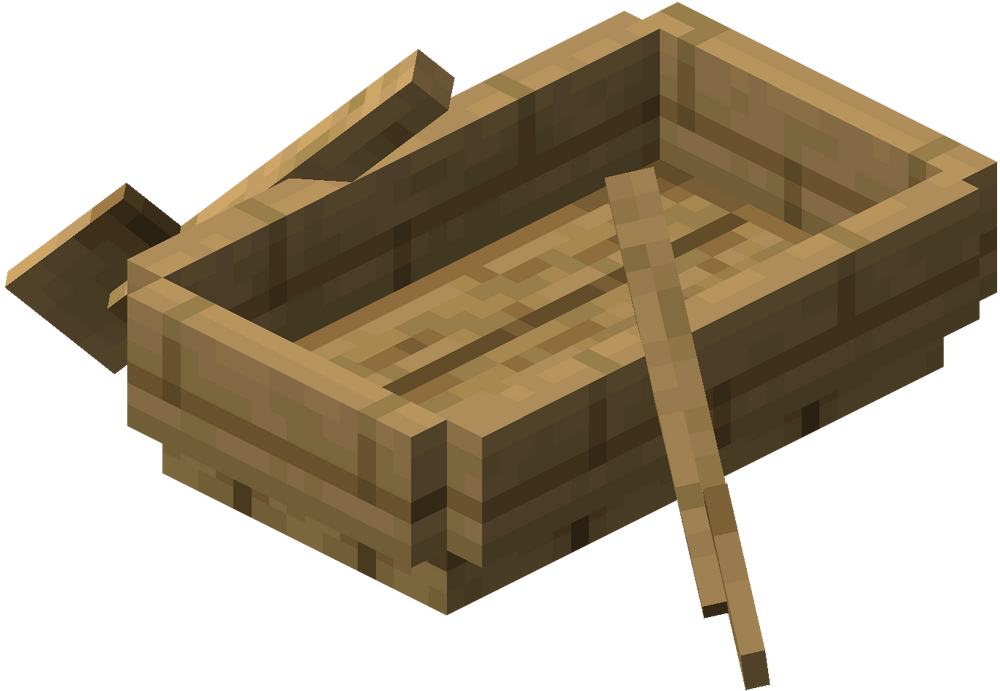}} \\
        \hline
        
        \multirow{10}{*}{Medium} 
        & 11 & Find a \underline{tree} \raisebox{-0.3\height}{\includegraphics[width=0.5cm]{figures/minecraft_icons/jungle_tree.png}} in the \underline{forest} \raisebox{-0.3\height}{\includegraphics[width=0.5cm]{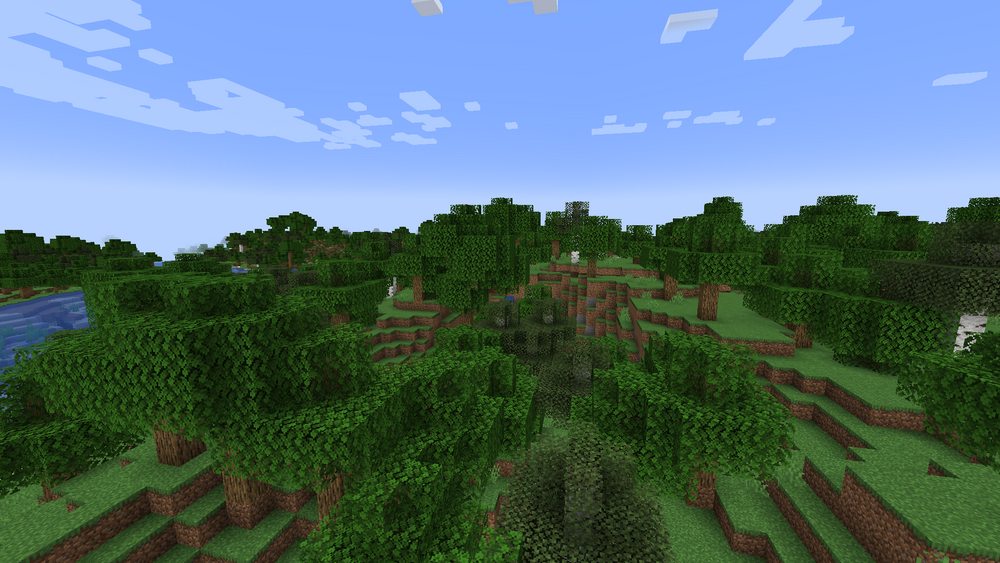}} \\ 
        & 12 & Find a \underline{pig} \raisebox{-0.3\height}{\includegraphics[width=0.5cm]{figures/minecraft_icons/pig.png}} on \underline{grass} \raisebox{-0.3\height}{\includegraphics[width=0.5cm]{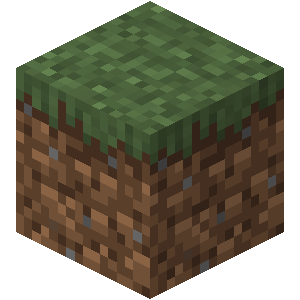}} \\ 
        & 13 & Find a \underline{cow} \raisebox{-0.3\height}{\includegraphics[width=0.5cm]{figures/minecraft_icons/cow.png}} in the \underline{desert} \raisebox{-0.3\height}{\includegraphics[width=0.5cm]{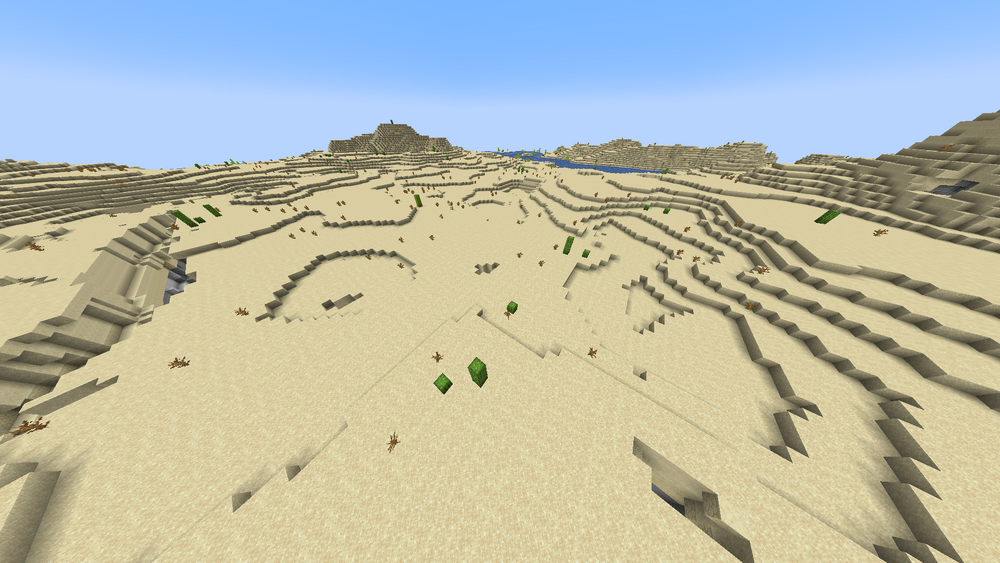}} \\ 
        & 14 & Craft a \underline{wooden sword} \raisebox{-0.3\height}{\includegraphics[width=0.5cm]{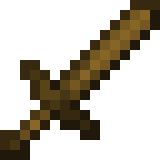}} \\ 
        & 15 & Craft a \underline{wooden pickaxe} \raisebox{-0.3\height}{\includegraphics[width=0.5cm]{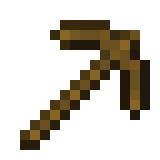}} \\ 
        & 16 & Craft a \underline{stone pickaxe} \raisebox{-0.3\height}{\includegraphics[width=0.5cm]{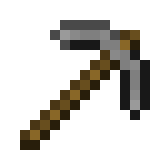}} \\ 
        & 17 & Smelt an \underline{iron ingot} \raisebox{-0.3\height}{\includegraphics[width=0.5cm]{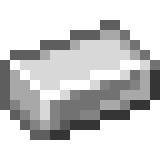}} \\ 
        & 18 & Smelt \underline{glass} \raisebox{-0.3\height}{\includegraphics[width=0.5cm]{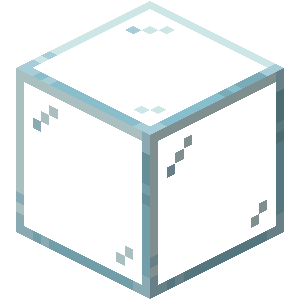}} \\ 
        & 19 & Cook \underline{beef} \raisebox{-0.3\height}{\includegraphics[width=0.5cm]{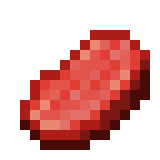}} \\ 
        & 20 & Cook \underline{mutton} \raisebox{-0.3\height}{\includegraphics[width=0.5cm]{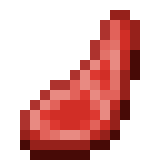}} \\ \hline
        
        \multirow{10}{*}{Hard} 
        & 21 & Find a \underline{pig} \raisebox{-0.3\height}{\includegraphics[width=0.5cm]{figures/minecraft_icons/pig.png}} near a \underline{grass} \raisebox{-0.3\height}{\includegraphics[width=0.5cm]{figures/minecraft_icons/grass.png}} in the \underline{forest} \raisebox{-0.3\height}{\includegraphics[width=0.5cm]{figures/minecraft_icons/forest.png}} during the \underline{daytime} \raisebox{-0.3\height}{\includegraphics[width=0.5cm]{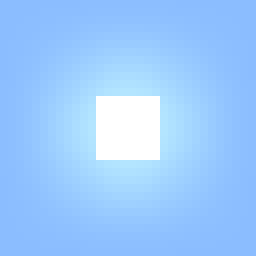}} \\ 
        & 22 & Find a \underline{cow} \raisebox{-0.3\height}{\includegraphics[width=0.5cm]{figures/minecraft_icons/cow.png}} in the \underline{desert} \raisebox{-0.3\height}{\includegraphics[width=0.5cm]{figures/minecraft_icons/desert.png}} during the \underline{daytime} \raisebox{-0.3\height}{\includegraphics[width=0.5cm]{figures/minecraft_icons/sun.png}} \\ 
        & 23 & Find a \underline{grass} \raisebox{-0.3\height}{\includegraphics[width=0.5cm]{figures/minecraft_icons/grass.png}} near a \underline{pig} \raisebox{-0.3\height}{\includegraphics[width=0.5cm]{figures/minecraft_icons/pig.png}} in the \underline{forest} \raisebox{-0.3\height}{\includegraphics[width=0.5cm]{figures/minecraft_icons/forest.png}} \\ 
        & 24 & Find a \underline{pig} \raisebox{-0.3\height}{\includegraphics[width=0.5cm]{figures/minecraft_icons/pig.png}} while wearing an \underline{iron helmet} \raisebox{-0.3\height}{\includegraphics[width=0.5cm]{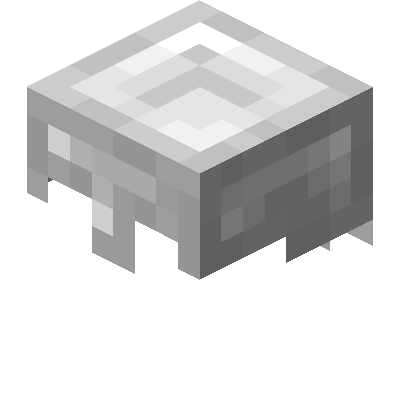}} \\ 
        & 25 & Craft an \underline{iron door} \raisebox{-0.3\height}{\includegraphics[width=0.5cm]{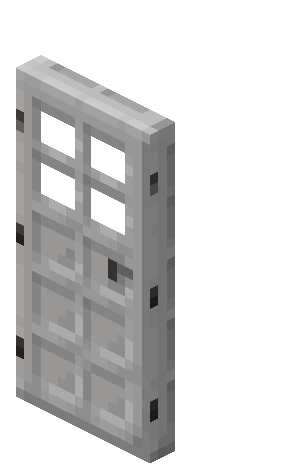}} \\ 
        & 26 & Craft an \underline{iron pickaxe} \raisebox{-0.3\height}{\includegraphics[width=0.5cm]{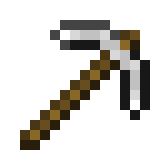}} \\ 
        & 27 & Craft an \underline{iron sword} \raisebox{-0.3\height}{\includegraphics[width=0.5cm]{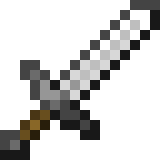}} \\ 
        & 28 & Craft a \underline{compass} \raisebox{-0.3\height}{\includegraphics[width=0.5cm]{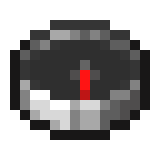}} \\ 
        & 29 & Kill a \underline{zombie} \raisebox{-0.3\height}{\includegraphics[width=0.5cm]{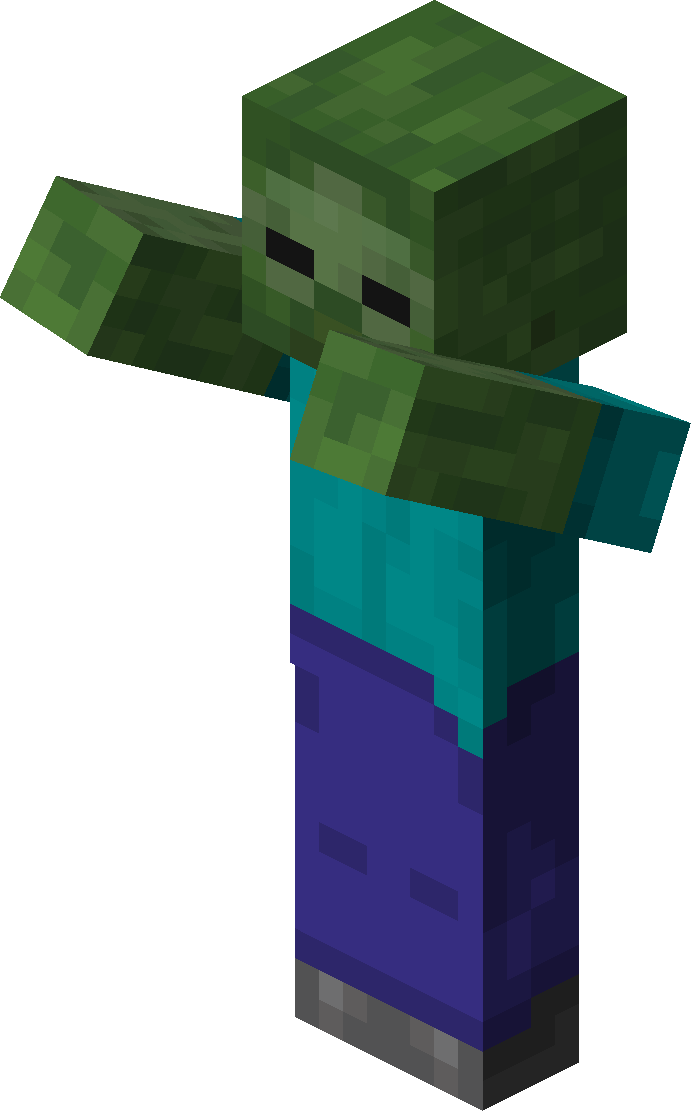}} with an \underline{iron sword} \raisebox{-0.3\height}{\includegraphics[width=0.5cm]{figures/minecraft_icons/iron_sword.png}} \\ 
        & 30 & Obtain a \underline{diamond} \raisebox{-0.3\height}{\includegraphics[width=0.5cm]{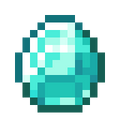}}\\ \hline
        
    \end{tabular}
    \caption{Full task details. 30 tasks evenly distributed as easy, medium, and hard. \underline{Underlines} label different information categories in Minecraft, highlighting how the complexity varies at each level.}
    \label{tab:task_settings}
\end{table}

\section{Detailed Model Descriptions}
\label{appendix:model_description}
\noindent \textbf{GPT-4V:} This vision-capable variant of GPT-4 excels at processing both visual and textual inputs, making it a powerful tool for tackling tasks within the visually complex Minecraft environment. Unlike its predecessors, GPT-4V's ability to seamlessly combine perception and reasoning allows for sophisticated decision-making and planning. The GPT-4 series has already demonstrated its efficacy in Minecraft-based research. For instance, Voyager \cite{wang2023voyager}, the first LLM-powered embodied lifelong learning agent, used GPT-4 to facilitate continuous exploration, skill acquisition, and task execution without human intervention. Voyager’s architecture included an automatic curriculum for exploration and a skill library to store and retrieve executable code, allowing agents to adapt and improve iteratively. Similarly, Optimus-1 \cite{li2024optimus} employs GPT-4V to refine its planning processes, focusing on logical reasoning and task generalization. These implementations underscore GPT-4V's pivotal role in advancing embodied AI research, offering exceptional capabilities for both exploration and problem-solving.

\noindent \textbf{MineLLM:} Tailored specifically for tasks within Minecraft, MineLLM represents a significant leap in AI development for complex embodied environments. As a central component of the MP5 framework \cite{qin2024mp5}, MineLLM is designed to tackle the unique challenges posed by Minecraft's open-ended tasks. It combines the image visual encoder from MineCLIP \cite{fan2022minedojo} with the Vicuna-13B-v1.5 language model for integrating visual perception with natural language understanding. Trained on a vast dataset of 500,000 Minecraft-specific image-text instruction pairs, MineLLM can generate detailed insights about the game environment, answer complex queries, and provide contextual guidance for planning and execution. Its integration into MP5 enables the framework to address context- and process-dependent tasks with remarkable success rates, achieving a 91\% success rate on context-dependent tasks and demonstrating exceptional adaptability in novel scenarios. 

\noindent \textbf{STEVE:} The STEVE series represents another advancement in language model-driven embodied agents for the Minecraft environment \cite{zhao2025see}. Built upon the foundation of LLaMA-2 \cite{touvron2023llama}, STEVE integrates powerful language capabilities tailored to enhance task reasoning, contextual understanding, and interaction. At its core, the language model in the STEVE series excels at decomposing complex objectives into actionable subtasks through iterative reasoning and hierarchical planning. This allows STEVE agents to process high-level instructions effectively and generate detailed plans for task execution. The STEVE series relies heavily on its ability to adapt to Minecraft-specific tasks. To this end,  \citet{zhao2025see} curated the STEVE-21K dataset, containing 20K knowledge-based question-answering pairs and 200+ skill-code pairs that directly enhance the model's contextual understanding and task reasoning. These adaptations enable the language model to seamlessly integrate with perception and action modules, driving coherent decision-making in real time. Furthermore, STEVE agents leverage advanced contextual awareness to refine their decision-making processes, significantly outperforming prior benchmarks in task decomposition and completion efficiency. The series also demonstrated up to 1.5x faster progression in complex tasks like unlocking tech trees and up to 2.5x quicker performance in block search scenarios compared to other state-of-the-art models.

\section{Additional Experiments}
\textbf{Do Hard Tasks Lead to Poor Calibration?}
We use the best-performing GPT-4V as our agent backbone and withhold any execution policies to reduce computation costs. We set the maximum episode length as 12,000 to provide enough coverage for all task difficulties. Results are shown in Table \ref{tab:task_difficulty_results}.

\begin{table}[h]
\centering
\begin{tabular}{c c c c c}

\textbf{Task} & \textbf{Policies} & \textbf{ECE (↓)} & \textbf{AUROC (↑)} & \textbf{Success Rate (↑)} \\  \hline
\multirow{5}{*}{\textbf{Easy}} 
    & Vanilla            & 0.26 & 0.76 & 84\% \\ 
    & Self-Intervention  & 0.26 & 0.76 & 92\% \\ 
    & CoT                & \textbf{0.11} & 0.78 & \textbf{94\%} \\ 
    & P\&S               & 0.12 & \textbf{0.80} & 82\% \\ 
    & Top-K              & 0.32 & 0.72 & 74\% \\ \hline
\multirow{5}{*}{\textbf{Medium}} 
    & Vanilla            & 0.35 & 0.54 & 52\% \\ 
    & Self-Intervention  & 0.35 & 0.51 & 44\% \\ 
    & CoT                & \textbf{0.22} & \textbf{0.58} & \textbf{54\%} \\
    & P\&S               & 0.22 & 0.55 & 48\% \\ 
    & Top-K              & 0.40 & 0.47 & 32\% \\ \hline
\multirow{5}{*}{\textbf{Hard}} 
    & Vanilla            & 0.33 & 0.58 & 17\% \\ 
    & Self-Intervention  & 0.35 & 0.52 & 12\% \\ 
    & CoT                & \textbf{0.31} & 0.68 & 18\% \\ 
    & P\&S               & 0.32 & \textbf{0.71} & \textbf{18\%} \\
    & Top-K              & 0.41 & 0.49 & 8\% \\ 
\end{tabular}
\vspace{-0.2cm}
\caption{ECE, AUROC, and Success Rates for Different Task Difficulties and Elicitation Policies. Lower ECE and higher AUROC/Success Rates indicate better performance.}
\label{tab:task_difficulty_results}
\vspace{-0.3cm}
\end{table}

For \textbf{Easy tasks}, CoT demonstrated the best performance, achieving the lowest ECE (0.11) and the highest success rate (94\%), followed by P\&S, which recorded the highest AUROC of 0.80 and a success rate of 82\%. Self-Intervention performed comparably in calibration (ECE = 0.26, AUROC = 0.76). Top-K underperformed, with the highest ECE (0.32) and the lowest success rate (74\%), indicating limitations in leveraging task simplicity. For \textbf{Medium tasks}, all policies showed noticeable declines in performance. CoT emerged as the best overall, with an ECE of 0.22, an AUROC of 0.58, and a success rate of 54\%, balancing calibration and task success effectively. P\&S followed closely with similar calibration (ECE = 0.22) but a slightly lower AUROC (0.55) and success rate (48\%). For \textbf{Hard tasks}, performance further degraded across all policies. CoT and P\&S maintained relative superiority, with CoT achieving an ECE of 0.31, AUROC of 0.68, and a success rate of 22\%, while P\&S recorded slightly worse calibration (ECE = 0.32) and the highest AUROC (0.71) but tied for a success rate of 18\%.

These results confirm our hypothesis that as task difficulty increases, confidence calibration significantly deteriorates, with the ECE gap increasing as high as 0.20 (Easy CoT vs. Hard CoT). However, the results also demonstrate that structured elicitation policies, such as CoT and P\&S, consistently prove effective in handling calibration, failure prediction, and task success across task difficulties. Additionally, simpler policies like Self-Intervention also show moderate success, particularly in easier tasks, suggesting their utility in less demanding scenarios.

\end{document}